\documentclass[10pt]{article}
\usepackage[preprint]{tmlr}

\usepackage{hyperref}
\usepackage{url}
\usepackage{booktabs}
\usepackage{longtable}
\usepackage{array}
\usepackage{calc}
\usepackage{graphicx}
\usepackage{amsmath}
\usepackage{amssymb}
\usepackage{textcomp}
\usepackage{etoolbox}
\usepackage{caption}
\newcounter{none}

\makeatletter
\patchcmd\longtable{\par}{\if@noskipsec\mbox{}\fi\par}{}{}
\makeatother

\AtBeginEnvironment{longtable}{\footnotesize}
\graphicspath{{./}}

\title{The Evaluation Protocol Determines the Result: An Independent Reproduction of LeWorldModel on TwoRoom}

\author{\name Joyjeet Singh \email joyjeetsingh1@gmail.com \\
      \addr Independent researcher \\ ORCID 0009-0005-1512-7439}

\begin{document}
\maketitle

\begin{abstract}
LeWorldModel trains a latent world model with a prediction loss and a single
anti-collapse regulariser, and reports approximately 87\% of goals reached on
TwoRoom --- its simplest diagnostic environment, where comparable methods report
97--100\%. We reproduce that result by independent reimplementation, on four
rented GPU-runs costing about six dollars each, with all evaluation on one
laptop CPU.

We reach \textbf{94.0\%} at the repository's evaluation goal offset, against \textbf{84.0\%}
for the authors' own released checkpoint measured under our protocol on
identical episodes, and we reproduce the reported representation result directly
(position probe Pearson r = 0.9988 against a reported 0.996). Reaching that
point required correcting \textbf{four conventions that determine the outcome and
appear in no released configuration file}: dense action gathering across a
frameskip block, a programmatically-set action-encoder width, ImageNet pixel
normalisation, and action z-scoring. A reproducer following the released
configurations alone obtains a model whose predictor cannot converge.

\textbf{The evaluation protocol is itself contested by the released material}, and we
report which reading reproduces. The paper's appendix and the repository's
evaluation configuration specify different goal offsets and step budgets; on the
authors' own released weights these yield 14.0\% and 84.0\% respectively, and only
the configuration's values reproduce the reported figure. On fifty identical
episodes, changing nothing but how the goal is constructed moves that same
checkpoint from 84.0\% to 8.0\%.

Two findings generalise beyond this reproduction. First, \textbf{one-step prediction
accuracy does not predict long-horizon planning success}: across three
checkpoints spanning a sevenfold range in prediction error --- including the
authors' own --- accuracy orders short-horizon success monotonically and fails to
order long-horizon success at all, where the two most accurate checkpoints
finish farther from the goal than a random-action policy. Second, \textbf{a batch
normalisation layer inflated our reported validation loss by a factor of up to 300}, concealing for three training runs a training loss that was flat
or descending throughout; we give the two conditions under which this occurs and
a cheap check for it.

We also report a pre-registered mechanism-level result that did not survive a
change of checkpoint, and what we take from that.

\begin{center}\rule{0.5\linewidth}{0.5pt}\end{center}
\end{abstract}

\section{1 Introduction}\label{introduction}

LeWorldModel \citep{maes2026lewm} proposes a latent world model trained with a
prediction loss and a single anti-collapse regulariser, deliberately without the
exponential moving averages, frozen encoders and auxiliary objectives that
comparable methods require. Among the environments it evaluates, TwoRoom is the
simplest: a point agent in a two-room arena joined by a single door, with an
action space of two dimensions and deterministic dynamics. It is also where the
method looks weakest. The paper reports approximately 87\% of goals reached
there, against 97--100\% for the baselines it compares against on the same task
\citep[Fig.~6]{maes2026lewm}. A method's behaviour on its easiest diagnostic is informative,
and an anomaly on that diagnostic is worth resolving before drawing conclusions
from harder ones.

We set out to reproduce three claims on TwoRoom: that the learned encoder
recovers agent position under a linear probe at a Pearson correlation of approximately 0.996, that
planning over the learned model reaches approximately 87\% of goals, and that the
released configuration produces such a model within the stated training budget.
We reimplemented the method from the released code and paper rather than
rerunning the released training script, trained four models on rented GPUs at
roughly six dollars each, and ran every evaluation on a single laptop CPU.

The reproduction succeeded, but not on the first attempt, and the failures along
the way turned out to be more useful than the success. Three of our four
training runs appeared not to converge. Every planning number we measured for
three of those runs was worthless. A carefully pre-registered mechanism-level
result did not survive a change of checkpoint. In each case the cause lay in our
instrumentation or in an undocumented convention rather than in the method under
study, and in each case we reported a confident and wrong conclusion before
finding it. This paper reports the reproduction and the failures together,
because the failures are what a reproducer attempting the same work would most
benefit from knowing.

\subsection{Contributions}\label{contributions}

\textbf{We reproduce the reported result and exceed it.} With four undocumented
pipeline conventions corrected and a normalisation artifact repaired, our
checkpoint reaches \textbf{94.0\%} of goals at the repository's evaluation goal
offset (\S{}4.5), against a reported \textasciitilde87\% and against \textbf{84.0\%} for the authors'
own released checkpoint measured under our protocol on identical episodes. The
representation result reproduces directly: $R^{2}$ 0.9977 under a linear probe
(\S{}4.1).

\textbf{We show that the released material publishes two evaluation protocols that
disagree, and that only one reproduces the reported figure} (\S{}4.2). Appendix
F.1 specifies a 150-step budget with the goal sampled 100 frames ahead; the
released evaluation configuration specifies 50 and 25. On the authors' own
released weights these yield 14.0\% and 84.0\%. On fifty identical episodes, with
identical weights and an identical planner, changing only the goal construction
moves that checkpoint from 84.0\% to 8.0\% (matched-pair p \textless{} 10\textsuperscript{-8}). We further
show that the appendix's protocol cannot be followed as written on the released
dataset: episode lengths cap at 101 frames, so a goal 100 frames ahead admits
exactly one legal start rather than the sampled one the appendix specifies, and
the eligible episodes are precisely the 6,056 in which the data-collection
policy ran out of time.

\textbf{We identify four conventions that determine the outcome and appear in no
configuration file} (\S{}3.2): actions must be gathered densely across a frameskip
block rather than sub-sampled, the action-encoder width is set programmatically
from that block, pixels are ImageNet-normalised, and actions are z-scored by
dataset statistics with NaN rows removed. A reproducer following the released
configurations alone obtains a model whose predictor cannot converge, and we
quantify why: the sub-sampled action implies a displacement wrong by roughly
twice the movement it is required to explain.

\textbf{We show that one-step prediction accuracy does not predict long-horizon
planning success} (\S{}5.3). Across three checkpoints spanning a sevenfold range
in one-step prediction error --- including the authors' own --- accuracy orders
short-horizon planning success monotonically and fails to order long-horizon
success at all. At the longer horizon the two most accurate checkpoints finish
\emph{farther} from the goal than a random-action policy does, and the least accurate
is the strongest planner by a wide margin.

\textbf{We document two measurement failures that a probe would not catch} (\S{}4.3,
\S{}5.1). A 32-pixel debugging fixture, visually near-indistinguishable from the
real environment, sat twenty-five times farther from the training distribution
than training frames sit from each other, and produced below-random planning
results for three paid runs while a position probe read $R^{2}$ 0.99 throughout. And
a batch-normalisation layer specified by the released configuration inflated our reported validation loss by a factor of up to 300, concealing a
training loss that was flat or descending throughout. We give the
conditions under which the second occurs --- a normalisation layer whose running
variance is far below its activation scale, combined with weights still moving
--- and show that the authors' released checkpoint does not satisfy the first
and is therefore unaffected.

\textbf{We report a pre-registered result that did not survive} (\S{}5.2). A same-room
planning advantage of +39.1 points at p = 3.4 $\times$ 10\textsuperscript{-8}, distance-matched
pair-by-pair with reachability verified at 100\% by an oracle and the episode
list committed before evaluation, falls to +12.7 points under a change of action
scaling and to $-$6.4 points on a different checkpoint. We report all three arms,
and take from it that effect sizes measured on a single reproduction checkpoint
should not be read as properties of the method.

\subsection{Scope}\label{scope}

All results concern the TwoRoom diagnostic. We make no claim about the original
paper's embodied or zero-shot results, about its other environments, or about
the method at scales other than the 18.03M-parameter configuration studied here.
Every figure in this paper comes from a single seed. Our full deviation set is
given in Table 1, our limitations in \S{}7, and all code, configurations,
evaluation reports and gate outputs are available at (github.com/joyjeet-singh/tinylab).

\section{2 Scope of reproducibility}\label{scope-of-reproducibility}

We test three claims the original makes about TwoRoom. Each is stated below in
the form the original makes it, followed by our verdict and a pointer to the
evidence. We also state two claims we deliberately do not test.

\subsection{Claim 1 --- the encoder recovers agent position}\label{claim-1-the-encoder-recovers-agent-position}

\medskip\noindent
The original reports that a linear probe recovers the agent's position from
the learned embedding at a Pearson correlation of approximately 0.996 \citep[Tab.~3, App.~F.2]{maes2026lewm}.
\medskip

\textbf{Reproduced.} On 4,000 held-out frames, a ridge probe fitted on 80\% recovers
position at \textbf{$R^{2}$ 0.9977} (Pearson r 0.9988), and a two-layer network on the
same split reaches 0.9994 (\S{}4.1). The result appears within a single training epoch and is robust
across every pipeline configuration we trained, including those whose predictor
does not converge.

We report one methodological caveat that bears on any comparison of probe
values. Our own per-epoch training logs report probe scores between 0.9305 and
0.9974 for the same encoders, a spread that vanishes to 0.0006 under a single protocol on identical frames. Probe values are protocol-dependent; we state ours
(4,000 held-out frames, ridge, 80/20 split) wherever we report one.

\subsection{Claim 2 --- planning over the learned model reaches approximately 87\%}\label{claim-2-planning-over-the-learned-model-reaches-approximately-87}

\medskip\noindent
The original reports approximately 87\% of goals reached under cross-entropy
method planning \citep[Fig.~6]{maes2026lewm}.
\medskip

\textbf{One of the two published evaluation protocols reproduces the figure; our
corrected checkpoint exceeds it; the comparison between checkpoints is not
established at our sample size.} Three measurements bear on this.

First, the authors' own released checkpoint, driven through our evaluation
harness with only the weights changed, reaches \textbf{42/50 = 84.0\%} under the
released repository's evaluation configuration --- a goal offset of 25 frames and
a 50-step budget. A one-sample test against 0.87 gives p = 0.53. Under the
protocol Appendix F.1 describes --- a goal 100 frames ahead and a 150-step budget
--- the same weights reach \textbf{14.0\%} (\S{}4.2). Our harness therefore recovers the
reported result from the reported weights under one published protocol and not
the other, and the released material does not tell us which was used.

Second, the goal construction alone accounts for most of that spread. On fifty
identical episodes, with identical weights, an identical planner and an
identical driving convention, changing only how the goal is defined moves their
checkpoint from 84.0\% to 8.0\% (matched-pair p \textless{} 10\textsuperscript{-8}, \S{}4.2). No property of any
model is involved in that difference.

Third, our own corrected checkpoint reaches \textbf{47/50 = 94.0\%} at the
repository's goal offset, with a 95\% interval of {[}83.8\%, 97.9\%{]} that contains
the reported figure (\S{}4.5). Against the authors' checkpoint on identical
episodes the difference is not established at our sample size (p = 0.0625), and
the reported 87\% was obtained under the authors' own episode selection, which
the released material does not describe in reproducible detail.

The conflict is therefore consequential rather than cosmetic, and we do not
resolve it. At the paper's goal offset our best checkpoint reaches 20.0\% within
the repository's budget and 26.0\% within the paper's; across the three
checkpoints we evaluate, the figure spans 12.0\% to 54.0\% at the shorter budget
and 14.0\% to 80.0\% at the longer (\S{}5.3). We report both offsets throughout and
quote no planning number without stating the offset and budget it was measured
under.

\subsection{Claim 3 --- the released configuration produces such a model in the stated budget}\label{claim-3-the-released-configuration-produces-such-a-model-in-the-stated-budget}

\medskip\noindent
The paper's appendix states ten training epochs \citep[App.~E]{maes2026lewm}; the released
repository configuration specifies one hundred (the released \texttt{le-wm} repository configuration).
\medskip

\textbf{Not reproduced as released; reproduced once four undocumented conventions are
corrected.} Our reimplementation of the released configuration plateaus: its
training loss reaches \textasciitilde0.30 within the first epoch and moves less than 4\% over
the following nine (\S{}4.4). With dense action gathering, the programmatic action
encoder width, ImageNet pixel normalisation and action z-scoring applied --- none
of which appears in any released configuration file --- the training loss descends
monotonically to a held-out value 36 times lower, within the same ten epochs
(\S{}3.2, \S{}4.4).

We are explicit that this is a statement about our reimplementation. We did not
rerun the authors' training script, and we make no claim that their training
procedure fails. Our verdict is that the released \emph{configuration}, as
implementable from the released \emph{configuration files}, is insufficient to
specify a converging run.

\subsection{Not tested}\label{not-tested}

\textbf{The original's other environments and its embodied and zero-shot results.}
Our budget covered four training runs on one task. We make no claim about any of
these.

\textbf{Seed variance.} Every figure in this paper comes from a single seed. Our
evaluations are deterministic --- re-running one from the committed commit
reproduces its per-episode outcomes exactly --- so the variance we have not
measured is between-seed, not within-run. Where we report a difference between
checkpoints, we report its test and, where it is not established at our sample
size, say so.

\section{3 Method}\label{method}

\subsection{3.1 Model and objective}\label{model-and-objective}

We reimplement the architecture the released configuration specifies. The
encoder is a ViT-Tiny at 224 pixels with patch size 14, twelve layers and three
heads, producing a 192-dimensional embedding. A projector maps that embedding
through a one-layer MLP with batch normalisation, which the original states is
necessary because the encoder's final layer normalisation would otherwise
prevent its anti-collapse objective from being optimised effectively
\citep[\S3.1]{maes2026lewm}; the predictor is a
six-layer transformer with sixteen heads, head dimension 64 and MLP width 2048,
consuming a context of three frames; a second projection of the same shape as
the first is applied to the predictor's output. An action embedder maps each
step's action into the predictor's dimension. Our implementation totals
\textbf{18,034,670 parameters} against \textbf{18,034,590} for the reference checkpoint
reconstructed from its released configuration; the difference of 80 is exactly
the width of the action encoder's first layer, which the reference sets
programmatically (\S{}3.2).

The objective is the sum of a prediction term and a regularisation term. The
prediction term is the mean squared error between the predicted next embedding
and the encoded next frame, \textbf{with no stop-gradient on the target}: gradient
flows into both sides, which makes representational collapse an available
solution and is why the second term carries real weight. The regulariser is a
sketched isotropy test --- the cloud of embeddings is projected onto many random
directions and each one-dimensional shadow is tested against a standard normal
using an Epps--Pulley statistic --- computed per timestep across the batch and
averaged. We follow the released weighting of 0.09 with 17 quadrature knots and
1024 projections.

We reimplemented rather than reran. That choice is what surfaced the four
undocumented conventions of \S{}3.2: a rerun would have inherited them silently,
and a reproduction that inherits an undocumented convention has not tested
whether the release specifies it.

\subsection{3.2 Fidelity of the reimplementation}\label{fidelity-of-the-reimplementation}

A reproduction is only as trustworthy as its account of where it differs from
the original. We therefore audited our reimplementation element by element
against the reference \textbf{source}, not against its configuration files, and
recorded each element as matching, deviating, or unverified. The distinction
turned out to be decisive: of the four deviations that mattered most, none
appears in any configuration file, and all four sit in code that a reader
following the released configs would never open.

24 of 40 audited elements match exactly, including the encoder
architecture (ViT-Tiny/14 at 224 pixels), the predictor geometry, batch size,
weight decay, gradient clipping, and the SIGReg parameters. Our parameter count
of 18,034,670 differs from the reference checkpoint's only by the width of the
action encoder.

The four deviations were these. First, the reference gathers actions at full
rate and reshapes them to \texttt{(history\_len,\ frameskip\ $\times$\ action\_dim)} \citep[\texttt{stable\_worldmodel/data/buffer.py}]{maes2026swm},
whereas we sub-sampled one action per clip step; the released \texttt{config.json}
records an action-encoder width of ten, which is exactly frameskip five times
action dimension two. Second, and consequently, the action-encoder width is set
programmatically at training time (the \texttt{le-wm} code release, \texttt{train.py}) rather than in the config. Third,
pixels are ImageNet-normalised before resizing (the \texttt{le-wm} code release, \texttt{utils.py}); we divided by 255 and
stopped. Fourth, non-pixel columns are z-scored using dataset statistics with
NaN rows dropped (the \texttt{le-wm} code release, \texttt{train.py} and \texttt{utils.py}); we used raw actions, and the dataset contains exactly
one NaN action per episode, at the final step.

Two of these are independently corroborated by the released artifact rather than
only by the code: the checkpoint's action encoder has ten input channels, and
the same checkpoint attains a one-step prediction error of 0.410 relative to a
frozen-world baseline under ImageNet normalisation against 5.1 under raw
\texttt{{[}0,1{]}} inputs --- a twelvefold difference that makes the convention effectively
mandatory.

The action deviation is the most consequential and the easiest to quantify.
Because the environment is deterministic, the displacement across a clip step
equals the speed times the summed actions of that block. Measured on the
released dataset, the sub-sampled convention leaves a median error of 25.59
units against a typical per-block displacement of 13.3 --- the action supplied to
the predictor was wrong by roughly twice the movement it was required to
explain. The remaining deviations, all deliberate, are listed in Table 1.

\medskip\noindent
\textbf{Table 1: Fidelity of our reimplementation against the reference.} Reference
values are cited to implementation source rather than to configuration files;
the four rows marked \textbf{undocumented} are determined in code and appear in no
released configuration. ``Corrected'' indicates a deviation present in our
earlier runs and fixed in the final configuration (\S{}3.2). Source codes:
\textbf{B} = \texttt{stable\_worldmodel/data/buffer.py}; \textbf{T} = \texttt{le-wm/train.py};
\textbf{U} = \texttt{le-wm/utils.py}; \textbf{C} = the released \texttt{config.json}; \textbf{Y} = the
repository YAML configuration; \textbf{package} = the installed distribution.
A number after a colon is a line number in the version we audited, whose
commit is recorded in the repository.
\medskip

\subsubsection{Data pipeline}\label{data-pipeline}

{\def\LTcaptype{none} 
\begin{longtable}[]{@{}
  >{\raggedright\arraybackslash}p{(\linewidth - 8\tabcolsep) * \real{0.2000}}
  >{\raggedright\arraybackslash}p{(\linewidth - 8\tabcolsep) * \real{0.2000}}
  >{\raggedright\arraybackslash}p{(\linewidth - 8\tabcolsep) * \real{0.2000}}
  >{\raggedright\arraybackslash}p{(\linewidth - 8\tabcolsep) * \real{0.2000}}
  >{\raggedright\arraybackslash}p{(\linewidth - 8\tabcolsep) * \real{0.2000}}@{}}
\toprule\noalign{}
\begin{minipage}[b]{\linewidth}\raggedright
element
\end{minipage} & \begin{minipage}[b]{\linewidth}\raggedright
reference
\end{minipage} & \begin{minipage}[b]{\linewidth}\raggedright
source
\end{minipage} & \begin{minipage}[b]{\linewidth}\raggedright
ours
\end{minipage} & \begin{minipage}[b]{\linewidth}\raggedright
status
\end{minipage} \\
\midrule\noalign{}
\endhead
\bottomrule\noalign{}
\endlastfoot
clip frame indices & \texttt{base\ +\ arange(history\_len)\ $\cdot$\ frameskip} & B \texttt{\_gather\_clip} & identical & match \\
\textbf{action gathering} & dense: \texttt{base\ +\ arange(history\_len\ $\cdot$\ frameskip)}, reshaped to (T, frameskip $\times$ action\_dim) & B \texttt{\_gather\_clip} & sub-sampled one action per clip step & \textbf{undocumented --- corrected} \\
\textbf{action-encoder width} & set programmatically to frameskip $\times$ action\_dim = 10 & T:68, C & 2 & \textbf{undocumented --- corrected} \\
\textbf{pixel preprocessing} & ImageNet mean/std normalisation, then resize & U:6 & divide by 255 only & \textbf{undocumented --- corrected} \\
\textbf{action normalisation} & per-dimension z-score from dataset statistics, NaN rows dropped & T:65, U:25--32 & raw actions & \textbf{undocumented --- corrected} \\
frameskip & 5 & Y & 5 & match \\
train/validation split & 0.9 / 0.1 & T:74 & 0.9 / 0.1, separate data seed & match (mechanism differs) \\
\end{longtable}
}

\subsubsection{Architecture}\label{architecture}

{\def\LTcaptype{none} 
\begin{longtable}[]{@{}
  >{\raggedright\arraybackslash}p{(\linewidth - 8\tabcolsep) * \real{0.2000}}
  >{\raggedright\arraybackslash}p{(\linewidth - 8\tabcolsep) * \real{0.2000}}
  >{\raggedright\arraybackslash}p{(\linewidth - 8\tabcolsep) * \real{0.2000}}
  >{\raggedright\arraybackslash}p{(\linewidth - 8\tabcolsep) * \real{0.2000}}
  >{\raggedright\arraybackslash}p{(\linewidth - 8\tabcolsep) * \real{0.2000}}@{}}
\toprule\noalign{}
\begin{minipage}[b]{\linewidth}\raggedright
element
\end{minipage} & \begin{minipage}[b]{\linewidth}\raggedright
reference
\end{minipage} & \begin{minipage}[b]{\linewidth}\raggedright
source
\end{minipage} & \begin{minipage}[b]{\linewidth}\raggedright
ours
\end{minipage} & \begin{minipage}[b]{\linewidth}\raggedright
status
\end{minipage} \\
\midrule\noalign{}
\endhead
\bottomrule\noalign{}
\endlastfoot
encoder & ViT-Tiny, patch 14, 224 px & Y, C & identical & match \\
encoder depth / heads & 12 / 3 & C & 12 / 3 & match \\
embedding dimension & 192 & Y, C & 192 & match \\
predictor depth / heads & 6 / 16 & C & 6 / 16 & match \\
predictor head dim / MLP & 64 / 2048 & C & 64 / 2048 & match \\
projector hidden width & 2048, BatchNorm1d & C & 2048, BatchNorm1d & match \\
dropout & 0.1 & C & 0.1 & match \\
context frames (\texttt{history\_size}) & 3 & Y, C & 3 (Run 0, phase2); \textbf{1} (Runs 1--2) & deviation, Runs 1--2 \\
parameter count & 18,034,590 (rebuilt from C) & C & 18,034,670 & +80 = action encoder width \\
\end{longtable}
}

\subsubsection{Objective}\label{objective}

{\def\LTcaptype{none} 
\begin{longtable}[]{@{}
  >{\raggedright\arraybackslash}p{(\linewidth - 8\tabcolsep) * \real{0.2000}}
  >{\raggedright\arraybackslash}p{(\linewidth - 8\tabcolsep) * \real{0.2000}}
  >{\raggedright\arraybackslash}p{(\linewidth - 8\tabcolsep) * \real{0.2000}}
  >{\raggedright\arraybackslash}p{(\linewidth - 8\tabcolsep) * \real{0.2000}}
  >{\raggedright\arraybackslash}p{(\linewidth - 8\tabcolsep) * \real{0.2000}}@{}}
\toprule\noalign{}
\begin{minipage}[b]{\linewidth}\raggedright
element
\end{minipage} & \begin{minipage}[b]{\linewidth}\raggedright
reference
\end{minipage} & \begin{minipage}[b]{\linewidth}\raggedright
source
\end{minipage} & \begin{minipage}[b]{\linewidth}\raggedright
ours
\end{minipage} & \begin{minipage}[b]{\linewidth}\raggedright
status
\end{minipage} \\
\midrule\noalign{}
\endhead
\bottomrule\noalign{}
\endlastfoot
prediction loss & MSE, target \textbf{not} detached & T:39 & identical & match \\
total loss & prediction + $\lambda$ $\cdot$ regulariser & T:41 & identical & match \\
regulariser weight $\lambda$ & \textbf{0.09 in the configuration; \S{}3.1 and Alg. 1 state 0.1} & Y, \S{}3.1 & 0.09 (Run 0, phase2); \textbf{0.045} (Runs 1--2) & \textbf{conflict}; deviation, Runs 1--2 \\
regulariser knots / projections & 17 / 1024 & Y & 17 / 1024 & match \\
regulariser axis & per timestep, across batch & T & identical & match \\
\end{longtable}
}

\subsubsection{Optimisation}\label{optimisation}

{\def\LTcaptype{none} 
\begin{longtable}[]{@{}
  >{\raggedright\arraybackslash}p{(\linewidth - 8\tabcolsep) * \real{0.2000}}
  >{\raggedright\arraybackslash}p{(\linewidth - 8\tabcolsep) * \real{0.2000}}
  >{\raggedright\arraybackslash}p{(\linewidth - 8\tabcolsep) * \real{0.2000}}
  >{\raggedright\arraybackslash}p{(\linewidth - 8\tabcolsep) * \real{0.2000}}
  >{\raggedright\arraybackslash}p{(\linewidth - 8\tabcolsep) * \real{0.2000}}@{}}
\toprule\noalign{}
\begin{minipage}[b]{\linewidth}\raggedright
element
\end{minipage} & \begin{minipage}[b]{\linewidth}\raggedright
reference
\end{minipage} & \begin{minipage}[b]{\linewidth}\raggedright
source
\end{minipage} & \begin{minipage}[b]{\linewidth}\raggedright
ours
\end{minipage} & \begin{minipage}[b]{\linewidth}\raggedright
status
\end{minipage} \\
\midrule\noalign{}
\endhead
\bottomrule\noalign{}
\endlastfoot
optimiser & AdamW & Y & AdamW & match \\
learning rate & 5 $\times$ 10\textsuperscript{-5} & Y & 5 $\times$ 10\textsuperscript{-5} (Run 0, phase2); \textbf{1 $\times$ 10\textsuperscript{-5}} (Runs 1--2) & deviation, Runs 1--2 \\
weight decay & 1 $\times$ 10\textsuperscript{-3} & Y & 1 $\times$ 10\textsuperscript{-3} & match \\
batch size & 128 & Y & 128 & match \\
gradient clipping & 1.0 & Y & 1.0 & match \\
prediction steps & 1 & Y & 1 & match \\
learning-rate schedule & \textbf{none specified} & Y & none (Runs 0, 1, phase2); \textbf{cosine} (Run 2) & deviation, Run 2 \\
epochs & 100 & Y & 10 & deviation --- paper's appendix states 10 \\
precision & bfloat16 & Y & float32 & deviation --- benign \\
seed & 3072 & Y & 0 & deviation --- benign \\
\end{longtable}
}

\subsubsection{Evaluation}\label{evaluation}

{\def\LTcaptype{none} 
\begin{longtable}[]{@{}
  >{\raggedright\arraybackslash}p{(\linewidth - 8\tabcolsep) * \real{0.2000}}
  >{\raggedright\arraybackslash}p{(\linewidth - 8\tabcolsep) * \real{0.2000}}
  >{\raggedright\arraybackslash}p{(\linewidth - 8\tabcolsep) * \real{0.2000}}
  >{\raggedright\arraybackslash}p{(\linewidth - 8\tabcolsep) * \real{0.2000}}
  >{\raggedright\arraybackslash}p{(\linewidth - 8\tabcolsep) * \real{0.2000}}@{}}
\toprule\noalign{}
\begin{minipage}[b]{\linewidth}\raggedright
element
\end{minipage} & \begin{minipage}[b]{\linewidth}\raggedright
reference
\end{minipage} & \begin{minipage}[b]{\linewidth}\raggedright
source
\end{minipage} & \begin{minipage}[b]{\linewidth}\raggedright
ours
\end{minipage} & \begin{minipage}[b]{\linewidth}\raggedright
status
\end{minipage} \\
\midrule\noalign{}
\endhead
\bottomrule\noalign{}
\endlastfoot
environment & \texttt{swm/TwoRoom-v1} & package & identical, verified bit-level (\S{}3.3) & match \\
success criterion & registered env rule, distance \textless{} 16 & package & identical & match \\
CEM settings & 300 samples / 30 elite / variance 1.0; \textbf{30 iterations in the configuration, App. D states 10 for TwoRoom} & Y, App. D & both reported (\S{}4.2) & \textbf{conflict --- immaterial: 42/50 under either (\S{}4.2)} \\
horizon, action block & 5, 5 & Y & 5, 5 & match \\
step budget & \textbf{50 in the evaluation config; App. F.1 states 150} & Y, App. F.1 & both reported (\S{}4.2, \S{}5.3) & \textbf{conflict --- only the config's value reproduces (\S{}4.2)} \\
episodes evaluated & 50 & Y & 50 (220 for \S{}5.2) & match \\
goal offset & \textbf{25 in the evaluation config; App. F.1 states 100} & Y, App. F.1 & both reported (\S{}4.2, \S{}4.5, \S{}5.3) & \textbf{conflict --- only the config's value reproduces (\S{}4.2)} \\
episode selection & not published & --- & fixed random draw, seed 42; start at episode frame 0 & deviation --- unavoidable \\
receding horizon & 5; App. D states the entire optimised sequence is executed before replanning & Y, App. D & identical & match --- confirmed behaviourally (\S{}5.3) \\
\end{longtable}
}

\subsubsection{Environment}\label{environment}

{\def\LTcaptype{none} 
\begin{longtable}[]{@{}
  >{\raggedright\arraybackslash}p{(\linewidth - 4\tabcolsep) * \real{0.3333}}
  >{\raggedright\arraybackslash}p{(\linewidth - 4\tabcolsep) * \real{0.3333}}
  >{\raggedright\arraybackslash}p{(\linewidth - 4\tabcolsep) * \real{0.3333}}@{}}
\toprule\noalign{}
\begin{minipage}[b]{\linewidth}\raggedright
element
\end{minipage} & \begin{minipage}[b]{\linewidth}\raggedright
ours
\end{minipage} & \begin{minipage}[b]{\linewidth}\raggedright
note
\end{minipage} \\
\midrule\noalign{}
\endhead
\bottomrule\noalign{}
\endlastfoot
PyTorch & 2.2.2+cu121 & identical across all four runs \\
Python & 3.11.15 (Runs 0--2); \textbf{3.12.3} (phase2) & deviation --- benign, recorded \\
\end{longtable}
}

\medskip\noindent
\textbf{Table 1b: Our four training runs.} Runs 1 and 2 were exploratory and vary
three and four hyperparameters from the reference respectively; their results
are reported as ablations. Run 0 and phase2 differ \textbf{only} in the four
pipeline corrections, and form the controlled pair of \S{}4.4.
\medskip

{\def\LTcaptype{none} 
\begin{longtable}[]{@{}
  >{\raggedright\arraybackslash}p{(\linewidth - 8\tabcolsep) * \real{0.2000}}
  >{\raggedright\arraybackslash}p{(\linewidth - 8\tabcolsep) * \real{0.2000}}
  >{\raggedright\arraybackslash}p{(\linewidth - 8\tabcolsep) * \real{0.2000}}
  >{\raggedright\arraybackslash}p{(\linewidth - 8\tabcolsep) * \real{0.2000}}
  >{\raggedright\arraybackslash}p{(\linewidth - 8\tabcolsep) * \real{0.2000}}@{}}
\toprule\noalign{}
\begin{minipage}[b]{\linewidth}\raggedright
\end{minipage} & \begin{minipage}[b]{\linewidth}\raggedright
Run 0
\end{minipage} & \begin{minipage}[b]{\linewidth}\raggedright
Run 1
\end{minipage} & \begin{minipage}[b]{\linewidth}\raggedright
Run 2
\end{minipage} & \begin{minipage}[b]{\linewidth}\raggedright
phase2
\end{minipage} \\
\midrule\noalign{}
\endhead
\bottomrule\noalign{}
\endlastfoot
purpose & reference config, as reimplemented & exploratory bundle & Run 1 + schedule & corrected pipeline \\
learning rate & 5e-5 & 1e-5 & 1e-5 & 5e-5 \\
schedule & none & none & cosine\textsuperscript{1} & none \\
regulariser weight & 0.09 & 0.045 & 0.045 & 0.09 \\
context frames & 3 & 1 & 1 & 3 \\
action width & 2 & 2 & 2 & \textbf{10} \\
dense actions & no & no & no & \textbf{yes} \\
ImageNet pixels & no & no & no & \textbf{yes} \\
z-scored actions & no & no & no & \textbf{yes} \\
epochs & 10 & 10 & 10 & 10 \\
deviations from reference\textsuperscript{2} & 7 & 10 & 11 & \textbf{3} \\
used for & \S{}4.4, \S{}4.1 & ablation & \S{}4.5, \S{}5.2, \S{}5.3 planning & \S{}4.4, \S{}4.5, \S{}5.2, \S{}5.3 \\
\end{longtable}
}

\textsuperscript{1} A patch applied twice caused the scheduler to step twice per epoch, turning
the intended one-way cosine decay into a full cycle from 1 $\times$ 10\textsuperscript{-5} down to
1 $\times$ 10\textsuperscript{-7} and back. The incident is disclosed in \S{}3.4; the resulting learning-rate
sweep is what identified the normalisation artifact of \S{}4.3, and we report it
as an accident rather than a design.

\textsuperscript{2} Every element Table 1 marks as a deviation for that run, including the three
benign ones common to all four: ten epochs rather than one hundred, float32
rather than bfloat16, and seed 0 rather than 3072.

\subsection{3.3 Environment verification}\label{environment-verification}

Every planning number in this paper depends on the evaluation environment being
the same environment that generated the training data. We establish that
directly rather than assuming it.

Placing the agent at each recorded position and rendering gives a \textbf{pixel mean
absolute error of 0.00} against the corresponding recorded frame. Replaying the
recorded action sequence from a recorded state reproduces the recorded
trajectory with an error of \textbf{0.000} at one step and at forty. Encoding paired
real and re-rendered frames gives a median latent distance of \textbf{0.01}, against
a nearest-neighbour spacing within the real data of 2.43.

This last figure is the instrument we use as a precondition throughout. Every
evaluation in this paper computes it before planning and refuses to report a
success rate if it exceeds a threshold of 1.0. The measured values across all runs reported here lie between 0.004 and 0.014.

The check earns its place. An earlier phase of this work evaluated in a
32-pixel fixture built for cheap iteration on a laptop, where the same
instrument reads \textbf{61.03} --- twenty-five times the real data's own
nearest-neighbour spacing. Three training runs' worth of planning results were
produced there and are worthless. We describe that episode in \S{}5.1; here we note
only that the precondition exists because it was needed.

\subsection{3.4 Experimental protocol and gates}\label{experimental-protocol-and-gates}

Rented compute forces a discipline that is worth stating, because it shaped what
we were able to conclude. Our budget allowed four training runs. A run that
fails for a preventable reason is not recoverable, so we adopted a rule that a
run counts against the budget only after a set of executable checks passes.

Four gates ran before each launch. \textbf{G1} clones the repository at the committed
head into a temporary directory, resolves every local module the training entry
point imports transitively, and verifies that all of them, and the configuration,
are present in the clone and compile there. This catches the classic failure of
a run that works locally because it imports a file that was never committed.
\textbf{G2} compares every element of the training configuration against
source-derived reference values, asserts the loader's data contract --- including
the physics identity of \S{}3.2 --- builds the model, and runs a short CPU training
loop checking for finite losses and for the regulariser being wired into the
objective. A deviation does not fail this gate, but an \emph{unexplained} deviation
does: each must be listed with a reason. \textbf{G3} is the domain precondition of
\S{}3.3, embedded in every evaluation rather than run separately. \textbf{G4} is a
written statement, committed before launch, of what each possible outcome will
and will not license.

Two properties of this arrangement mattered more than we expected. Gates that
fail loudly are worth more than gates that are correct: of the gate failures we
investigated, more were caused by defects in the gate than in the run (\S{}6.2),
and each of those defects was itself a finding about what we had assumed. And
committing G4 before launch prevented at least one post-hoc reinterpretation: our
recorded prediction for the run of \S{}4.4 was wrong, and having written it down
made that unambiguous.

We also pre-registered one experiment in full --- the design, the decision rule for
every outcome, and the exact episode list --- before evaluating any of it (\S{}5.2).
Its outcome is reported in \S{}5.2, including the fact that the registered effect
did not survive subsequent analysis on other checkpoints.

All gate outputs, the pre-registration, and the expected-outcome statements are
in the repository.

\subsection{3.5 Computational requirements}\label{computational-requirements}

Four training runs on a single rented GPU, at approximately six US dollars each,
totalling about twenty-four dollars of compute. Each run is ten epochs over
roughly 780,000 clips at 224-pixel resolution and completes in a few hours.

Everything else ran on one laptop CPU with 8 GB of memory: all evaluation, all
planning, every probe, every gate, the environment verification, the fidelity
audit, and the normalisation recalibration of \S{}4.3. A planning evaluation of
fifty episodes takes fifteen to sixty minutes depending on the goal offset and
the model; the 220-episode matched-pair experiment of \S{}5.2 takes approximately
four hours. The recalibration procedure takes a few minutes.

Two consequences of this budget bear on our conclusions. First, we train for ten
epochs, following the paper's appendix, where the released repository
configuration specifies one hundred (Table 1); a hundred-epoch run was outside
our means, so our convergence result is a statement about the paper's stated
budget and not about the asymptote. Second, we have one seed per configuration,
and the differences we report between checkpoints are correspondingly qualified
(\S{}7).

We note the ratio deliberately. Twenty-four dollars of GPU time produced four
checkpoints; several hundred hours of CPU time produced everything that made
those checkpoints interpretable, including all four of the findings we consider
most transferable. A reproduction of this kind is not principally a compute
problem.

\section{4 Reproduction results}\label{reproduction-results}

\subsection{4.1 The representation reproduces, and is not the bottleneck}\label{the-representation-reproduces-and-is-not-the-bottleneck}

The original reports that a linear probe recovers agent position from the
learned embedding at a Pearson correlation of approximately 0.996 \citep[Tab.~3, App.~F.2]{maes2026lewm}. We
reproduce this. On 4,000 held-out frames drawn uniformly from the released
dataset, a ridge probe fitted on 80\% and evaluated on the remaining 20\%
recovers position at \textbf{$R^{2}$ 0.9977}, a Pearson correlation of 0.9988; a two-layer
MLP probe on the same split reaches $R^{2}$ 0.9994, confirming the linear probe is
not limited by its own capacity. We report $R^{2}$ throughout, since it is the
quantity our tooling computes, and give the correlation wherever a comparison
with the original requires it.

The protocol matters more here than the number. Our own training logs report
per-epoch probe values ranging from 0.9922 to 0.9974 for the reference-faithful
run and from 0.9305 to 0.9525 for the corrected-pipeline run --- an apparent
five-point difference between the two pipelines. Measured under a single
protocol on identical frames, with both checkpoints normalisation-repaired
(\S{}4.3), that difference is \textbf{0.0006} (0.9977 against 0.9971) and disappears
entirely under the non-linear probe (0.9994 against 0.9994). The in-training
probe fits far fewer samples, where ridge regularisation dominates at 192
dimensions. We report the common-protocol numbers throughout and recommend that
probe protocols be stated wherever probe values are compared, including within a
single paper's own logs.

A second measurement matters more for what follows. From a pair of consecutive
embeddings (z\_t, z\_\{t+k\}), the summed action executed between them is linearly
decodable at \textbf{$R^{2}$ 0.9290}; from their difference alone, at 0.8982. Together
with the position result this characterises the latent space precisely: it is a
near-linear encoding of agent position, and transitions within it carry the
action that produced them in linearly accessible form.

This has a direct consequence for the training results in \S{}4.3 and \S{}4.4. The
environment's dynamics are deterministic and, in position space, affine in the
action: displacement equals a fixed speed times the summed actions of the block
(\S{}3.3). Because the embedding is a near-linear encoding of position, the
forward map a predictor must learn in latent space is approximately as simple as
the true dynamics, and every quantity it requires is present and linearly
accessible in its inputs. \textbf{The failure to converge documented below is
therefore not an information-theoretic limitation of the representation. It is a
property of the predictor and its optimisation.}

The original's authors reach the same conclusion from their own results. The
caption to their probing table observes that although the method underperforms
PLDM in downstream planning on this environment, it matches or exceeds PLDM
across the probing metrics, and suggests that the planning gap is therefore not
due to a less informative representation but to the dynamics model or the
planning procedure itself \citep[Tab.~3 caption]{maes2026lewm}. Our measurements support that reading and
sharpen it: not only is the position information present, so is the action
information a planner needs, and the forward map is no harder in latent space
than in the true state space. \S{}5.3 takes up what happens when the planner is
given a predictor that is nonetheless far more accurate.

Finally, the regulariser does its job. Mean embedding spread remained within
0.830--0.960 across the reference-faithful run and 0.797--1.039 across the
corrected-pipeline run, with no monotone decline in either (Figure 1c). We
observed no representation collapse under any configuration we trained,
including at the reference learning rate where the prediction loss does not
settle. This supports the original's central architectural claim --- that the
two-term objective is sufficient to prevent collapse without an exponential
moving average, a frozen encoder, or auxiliary supervision --- independently of
whether the predictor converges.

\medskip\noindent
\textbf{Table 2: Encoder comparison under a single probe protocol.} 4,000 held-out
frames, identical for both models; each encoder receives the pixel convention
it was trained with. Ridge probes, 80/20 split. Both checkpoints are
recalibrated (\S{}4.3); measured before that repair, both read differently (\S{}5.4).
\medskip

{\def\LTcaptype{none} 
\begin{longtable}[]{@{}
  >{\raggedright\arraybackslash}p{(\linewidth - 4\tabcolsep) * \real{0.3333}}
  >{\raggedright\arraybackslash}p{(\linewidth - 4\tabcolsep) * \real{0.3333}}
  >{\raggedright\arraybackslash}p{(\linewidth - 4\tabcolsep) * \real{0.3333}}@{}}
\toprule\noalign{}
\begin{minipage}[b]{\linewidth}\raggedright
measurement
\end{minipage} & \begin{minipage}[b]{\linewidth}\raggedright
reference-faithful (Run 0)
\end{minipage} & \begin{minipage}[b]{\linewidth}\raggedright
corrected pipeline (phase2)
\end{minipage} \\
\midrule\noalign{}
\endhead
\bottomrule\noalign{}
\endlastfoot
position, linear probe & \textbf{0.9977} & 0.9971 \\
position, MLP probe & 0.9994 & 0.9994 \\
summed action from (z\_t, z\_\{t+k\}) & \textbf{0.9290} & 0.9132 \\
summed action from z\_\{t+k\} $-$ z\_t & 0.8982 & 0.8844 \\
effective rank (of 192) & 18.6 & 67.8 \\
mean embedding spread & 1.004 & 1.009 \\
\end{longtable}
}

\emph{Effective rank is discussed in \S{}5.4; it is listed here so the comparison is
presented once.}

\begin{center}
\includegraphics[width=\textwidth]{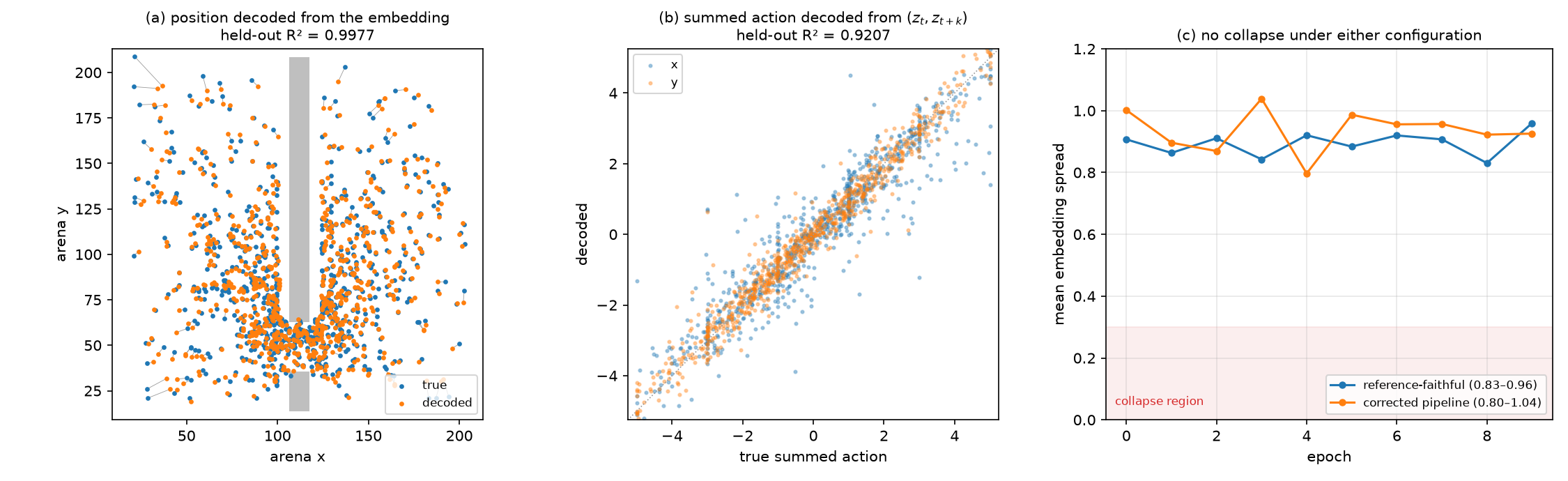}
\end{center}

\medskip\noindent
\textbf{Figure 1: The representation carries what the predictor needs.}
\textbf{(a)} Position decoded by a ridge probe against true position, in arena
coordinates, for 800 held-out frames; the dividing wall and door are drawn for
reference. Held-out $R^{2}$ = 0.9977. \textbf{(b)} Summed action decoded from a pair of
consecutive embeddings against the true summed action; held-out $R^{2}$ = 0.9207.
\textbf{(c)} Mean embedding spread per epoch for both training configurations; no
monotone decline appears in either, and no run collapsed. Panels (a) and (b)
use the reference-faithful checkpoint as trained; recalibrated, it probes at
0.9977 and 0.9290, and the corrected-pipeline checkpoint at 0.9971 and 0.9132
(Table 2, both recalibrated).
\medskip

\subsection{4.2 Two published evaluation protocols, and which one reproduces}\label{two-published-evaluation-protocols-and-which-one-reproduces}

A reproduction that reports a planning number is reporting the product of two
things: a model, and a protocol for evaluating it. If the protocol is wrong,
every number it produces is wrong in a way no amount of internal consistency
checking will reveal. Before reporting any figure from our own checkpoints, we
therefore ran the \textbf{authors' released checkpoint} --- not ours --- through our
evaluation harness, changing nothing but the weights.

Doing so required choosing a protocol, and the released material publishes two
that disagree. Appendix F.1 states that for TwoRoom the evaluation budget is 150
steps and the goal is sampled 100 timesteps in the future. The evaluation
configuration in the released repository uses a budget of 50 and a goal offset
of 25. We ran their checkpoint under both, and under five further readings
constructed from the paper's description of the task.

{\def\LTcaptype{none} 
\begin{longtable}[]{@{}
  >{\raggedright\arraybackslash}p{(\linewidth - 4\tabcolsep) * \real{0.3333}}
  >{\raggedright\arraybackslash}p{(\linewidth - 4\tabcolsep) * \real{0.3333}}
  >{\raggedright\arraybackslash}p{(\linewidth - 4\tabcolsep) * \real{0.3333}}@{}}
\toprule\noalign{}
\begin{minipage}[b]{\linewidth}\raggedright
goal construction
\end{minipage} & \begin{minipage}[b]{\linewidth}\raggedright
budget
\end{minipage} & \begin{minipage}[b]{\linewidth}\raggedright
success
\end{minipage} \\
\midrule\noalign{}
\endhead
\bottomrule\noalign{}
\endlastfoot
frame 25 later --- the released evaluation configuration & 50 & \textbf{42/50 = 84.0\%} \\
frame 100 later --- Appendix F.1 & 50 & 6/50 = 12.0\% \\
frame 100 later --- Appendix F.1 & 150 & 7/50 = 14.0\% \\
frame 100 later, action repeated rather than block-averaged & 150 & 5/50 = 10.0\% \\
frame 100 later, initial state sampled within the trajectory & 150 & 5/50 = 10.0\% \\
the episode's recorded target, episodes the data policy solved & 150 & 7/50 = 14.0\% \\
the episode's recorded target, all episodes & 150 & 4/50 = 8.0\% \\
\end{longtable}
}

\emph{Every figure in this table was read from a committed evaluation report; the
report path for each is listed in \texttt{docs/paper/results\_from\_disk.csv}. The fifth
row is discussed below: on this dataset that instruction has no effect at an
offset of 100, and the row is identical to the fourth for that reason.}

Only the first reproduces the reported figure. Under it the checkpoint reaches
42 of 50 goals; a one-sample test against 0.87 gives p = 0.53, and the 95\%
Wilson interval, {[}71.5\%, 91.7\%{]}, contains the reported figure. A random-action
control over the same episodes reaches 9 of 50. Under the protocol the paper
describes, the same weights reach 14.0\%.

The consequence is worth stating plainly, because it is the kind of thing a
reproduction exists to find: \textbf{the evaluation protocol described in the paper
does not reproduce the paper's reported result on the paper's own released
weights, and the released repository's evaluation defaults do.} We do not know
which the authors used, and we do not assume; \S{}6.4 records that we asked.

\subsubsection{The goal construction, not the model, determines the number}\label{the-goal-construction-not-the-model-determines-the-number}

The last row of the table draws the same fifty episodes as the first. Neither
imposes a minimum episode length, and both sample with the same seed, so the
two runs are a paired comparison on identical episodes, with identical weights,
an identical planner and an identical driving convention. They differ in how the goal is constructed and, because each protocol carries its own step budget, in budget: 50 steps for the first and 150 for the second. That difference runs against the comparison rather than for it --- the recorded-target arm was given 3 times as many steps and still collapsed. The result is 84.0\% against 8.0\%: 39 episodes are
solved under the first and missed under the second, one is solved under the
second alone, and a matched-pair test gives $\chi^{2}$ = 34.2, p \textless{} 10\textsuperscript{-8}. No property of
the model changed between those two numbers.

The failure mode differs as well as the rate. Under the recorded-target
construction the planner ends a mean of 133.4 units from its goal, having
started a mean of roughly 136 away --- no net progress --- and 29 of
the 50 episodes finish farther from the goal than they began. Under the offset-25
construction the same weights close from a mean of 46 units to 25.

\subsubsection{Why the long-horizon reading cannot be followed as written}\label{why-the-long-horizon-reading-cannot-be-followed-as-written}

That the offset-100 readings score poorly is a property of the released dataset
rather than of the planner, and it is visible without running a model at all.

The dataset contains 10,000 episodes of mean length 92.1 and maximum length
exactly 101. An episode can supply a goal 100 frames after its start only if it
runs longer than 100 frames, so the eligible set is precisely the 6,056
episodes that reached the length cap. That cap is a timeout. Episodes ending
before it do so because the data policy reached its target: all 3,944 of them
are recorded as successful, and they finish a mean of 13.6 units from the
target against a 16-unit success radius, whereas capped episodes finish a mean
of 73.1 units away. Of the 6,056 eligible episodes, only 91 --- 1.5\% --- were
solved by the policy that generated them. Evaluating at an offset of 100 therefore asks the planner to reach the point at which a failing policy ran out of time.

This argument holds under the reading that the offset is exact. \texttt{stable-worldmodel} describes the offline protocol as constraining the \emph{maximum} number of steps separating start and goal rather than fixing it, and under that reading all 10,000 episodes are eligible rather than 6,056, so the eligible set is no longer the data policy's failures and the argument above does not apply to it. We report the exact-offset reading because it is the one the appendix's own wording --- a goal sampled 100 timesteps in the future --- most directly supports, and because it is the reading under which the protocol is reproducible at all. We did not evaluate the alternative.

The same appendix specifies that the initial state is sampled from within the
trajectory rather than fixed at its first frame. On this dataset that
instruction cannot be followed at an offset of 100: an episode of exactly 101
frames whose goal lies 100 frames after its start admits exactly one legal
start, and the maximum sampling span across all 6,056 eligible episodes is
zero. At an offset of 25 the mean span is 66 frames and the instruction is
followed normally. We report this as a property of the released dataset rather
than as an error in the paper: the description is implementable at the offset
the configuration uses, and not at the offset the appendix states.

We draw one methodological point from this. Five hypotheses for the discrepancy
were tested and eliminated over a single day --- the action encoding, the context
length, the sampling of initial states, the step budget, and the number of
optimiser iterations. What identified the cause was none of them, but a
one-minute query against the distribution of episode lengths. When a
reproduction disagrees with its source, the data are worth interrogating before
the model (\S{}6.3).

\subsubsection{Two conventions the released material does not state}\label{two-conventions-the-released-material-does-not-state}

Reaching the 84.0\% figure required settling two things the released material
does not state, and both were settled by measurement rather than assumption.

\textbf{The input convention.} The released configuration specifies the architecture
but not the preprocessing. Encoding frames under raw \texttt{{[}0,1{]}} pixel values, their
checkpoint predicts the next latent state with an error 5.1 times a
frozen-world baseline --- five times \emph{worse} than assuming nothing moves. Under
ImageNet normalisation the same weights score 0.410. A twelvefold difference
makes the convention effectively mandatory, and it is discoverable only by
reading the reference's preprocessing chain or by measuring, as we did.

\textbf{The action convention.} Their action encoder accepts ten inputs while the
environment's action space has two dimensions. Reading the reference's clip
loader resolves this --- actions are gathered densely across a frameskip block and
concatenated, so ten is frameskip five times action dimension two (\S{}3.2). A
planner emitting one action held across a block corresponds to a block whose
mean is that action, and we supply it accordingly. Supplying the action
repeated rather than averaged is measurably equivalent on this checkpoint
(one-step error ratios 0.410 against 0.415), and the fourth row of the table
above shows it makes no material difference to planning either.

\subsubsection{An earlier failure, recorded}\label{an-earlier-failure-recorded}

We record one earlier failure here because it bears on how such numbers should
be read. Our first attempt at this calibration supplied the action at the wrong
scale, and produced \textbf{46.0\%} --- a number that arrived with every surrounding
check passing: the domain guard passed, the random-action control replicated
exactly, and the run completed without error. What identified it as an artifact
was not any check but the \emph{shape} of the failures: successes came unusually fast (median 5 steps against our checkpoint's 21) while misses were extreme overshoots, with 22 of 27 finishing farther from the goal than they started.
That is the signature of a planner whose model understates how far an action
moves the agent. Measuring the action scale directly then confirmed it, and the corrected figure is the 84.0\% above. The superseded run is committed rather than described: \texttt{exp\_wrongscale\_25/}, driven by a spec identical to the corrected one but for the two fields that were wrong. We report the episode as evidence that a
completed run with passing checks is not the same as a correct measurement, and
that a wrong answer of this kind is more readily caught by inspecting the
distribution of failures than by any single summary statistic.

\subsubsection{What this validation does and does not establish}\label{what-this-validation-does-and-does-not-establish}

Three properties are worth stating for what follows. First, it is independent of
everything we trained: no checkpoint of ours enters it, so the protocol used in
\S{}4.5 and \S{}5.3 is validated regardless of how our own training turned out.
Second, it is not a validation of our \emph{model} in any respect, and we do not use
it as one. Third, we verified separately that the authors' checkpoint requires
no normalisation recalibration --- its evaluation-to-training gap is 1.09$\times$ (\S{}4.3)
--- so the figures we report for it here and in \S{}5.3 are not affected by the
artifact described in that section.

Because the two published protocols disagree, we report our own checkpoints at
both goal offsets throughout, and we do not compare any single figure to the
reported 87\% without stating the offset and budget it was measured under.

\subsection{4.3 An evaluation-mode artifact concealed the training result}\label{an-evaluation-mode-artifact-concealed-the-training-result}

For three of our four training runs we recorded a per-epoch validation
prediction loss that oscillated by more than 100\% of its mean and showed no
trend across ten epochs. Read on its own, that series says the predictor does
not converge, and we reported it that way for a week. The per-step \textbf{training}
loss, written to the same log file, was flat or monotonically descending in
every one of those runs. We had not opened it.

The two series cannot both be describing the model's progress. To find out which
was misleading, we scored a single fixed checkpoint four ways: in evaluation
mode and in training mode, on held-out clips and on training clips. If the gap
were a generalisation gap, it would follow the data; if it were an artifact of
the evaluation procedure, it would follow the mode.

{\def\LTcaptype{none} 
\begin{longtable}[]{@{}
  >{\raggedright\arraybackslash}p{(\linewidth - 8\tabcolsep) * \real{0.2000}}
  >{\raggedright\arraybackslash}p{(\linewidth - 8\tabcolsep) * \real{0.2000}}
  >{\raggedright\arraybackslash}p{(\linewidth - 8\tabcolsep) * \real{0.2000}}
  >{\raggedright\arraybackslash}p{(\linewidth - 8\tabcolsep) * \real{0.2000}}
  >{\raggedright\arraybackslash}p{(\linewidth - 8\tabcolsep) * \real{0.2000}}@{}}
\toprule\noalign{}
\begin{minipage}[b]{\linewidth}\raggedright
checkpoint
\end{minipage} & \begin{minipage}[b]{\linewidth}\raggedright
eval mode, held-out
\end{minipage} & \begin{minipage}[b]{\linewidth}\raggedright
train mode, held-out
\end{minipage} & \begin{minipage}[b]{\linewidth}\raggedright
eval mode, train clips
\end{minipage} & \begin{minipage}[b]{\linewidth}\raggedright
train mode, train clips
\end{minipage} \\
\midrule\noalign{}
\endhead
\bottomrule\noalign{}
\endlastfoot
released configuration & 1.4585 & \textbf{0.3077} & 1.4791 & 0.2975 \\
corrected pipeline & 4.6034 & \textbf{0.0151} & 4.5525 & 0.0149 \\
\end{longtable}
}

The mode effect is +1.15 and +4.59. The data effect is +0.010 and \textbf{+0.0002}.
There is no generalisation gap in either run --- on the corrected checkpoint the
held-out and training losses are indistinguishable, the difference changing
sign between independent measurements of it at a magnitude of 0.0003 against a
loss of 0.015 --- and the training-mode held-out values match the training logs
to within 2\%. Training-mode figures are stochastic, dropout being active in
that mode, and vary in the third decimal between measurements.

The prediction loss is not the only quantity affected. The anti-collapse term is
distorted by the same mechanism and more severely: under the released
configuration it reads 107.1 in evaluation mode against 3.0 in training mode,
and under the corrected pipeline 156.5 against 1.4. A practitioner monitoring
that series in evaluation mode --- the natural thing to do, since it is the series
the evaluation loop writes --- would conclude that the regulariser was diverging
while it was in fact doing its job.

The mechanism is in the checkpoints, and it has two factors rather than one.
The projector's \texttt{BatchNorm1d}, specified by the released configuration
(the released \texttt{le-wm} repository configuration), carries a running variance of order 10\textsuperscript{-4} in all three of our
checkpoints. In evaluation mode the layer divides by the square root of that
quantity, so any drift between the stored statistics and the current activations
is amplified by a factor of 72 to 141; a squared error inflates that by two
further orders of magnitude. But amplification alone is not sufficient. The
checkpoint saved at the minimum of our accidental learning-rate cycle has an
amplification of 78 and a gap of exactly 1.00$\times$, because at a learning rate of
10\textsuperscript{-7} the weights had stopped moving and the running statistics had caught up.
The gap requires both a large amplification and weights that are still moving.

That this is a property of particular checkpoints rather than of the
architecture is established directly by the authors' released weights, which use
the same two \texttt{BatchNorm1d} layers. We measured their evaluation-to-training gap
on the same held-out clips and found \textbf{1.09$\times$} --- calibrated. Their projector's
running variance is 0.0172, \textbf{89 times larger than our corrected checkpoint's
0.00019}, so their evaluation mode divides by 0.131 where ours divides by
0.014, an amplification of 7.6 rather than 72. For contrast the second normalisation layer, \texttt{pred\_proj}, is not degenerate in
any of them: its running variance is 1.163 in the authors' checkpoint and 1.159
in our corrected one, and 0.844 and 0.202 in our two earlier runs --- in every
case orders of magnitude above the projector's. It is the projector alone that is near-degenerate in
ours.

Why our projector output is so much narrower than theirs we do not establish.
Training length is the obvious candidate --- we train for the ten epochs of the
paper's appendix against the repository's hundred (Table 1) --- but we have not
tested it, and we report it as open. The practical consequence does not depend
on the cause: their released checkpoint required no recalibration and the
figures we report for it in \S{}4.5 and \S{}5.3 are unaffected, while all three of
ours did.

Two further observations support this account.

First, the size of the gap tracks the learning rate. One of our runs applied a
cyclic schedule by accident (\S{}3.4), sweeping the rate from 1$\times$10\textsuperscript{-5} down to
1$\times$10\textsuperscript{-7} and back. The ratio of evaluation to training loss follows it with
r = +0.899 on a log-log scale, reaching exactly \textbf{1.00$\times$} at the minimum, where
the weights stop moving and the running statistics catch up, and reopening to
186$\times$ as the rate climbs again.

Second, the artifact is repairable without touching a weight. Resetting the
running statistics and accumulating a cumulative average over 100--200 training
batches in training mode --- a standard ``precise BN'' recalibration --- restores
agreement:

{\def\LTcaptype{none} 
\begin{longtable}[]{@{}
  >{\raggedright\arraybackslash}p{(\linewidth - 8\tabcolsep) * \real{0.2000}}
  >{\raggedright\arraybackslash}p{(\linewidth - 8\tabcolsep) * \real{0.2000}}
  >{\raggedright\arraybackslash}p{(\linewidth - 8\tabcolsep) * \real{0.2000}}
  >{\raggedright\arraybackslash}p{(\linewidth - 8\tabcolsep) * \real{0.2000}}
  >{\raggedright\arraybackslash}p{(\linewidth - 8\tabcolsep) * \real{0.2000}}@{}}
\toprule\noalign{}
\begin{minipage}[b]{\linewidth}\raggedright
checkpoint
\end{minipage} & \begin{minipage}[b]{\linewidth}\raggedright
eval before
\end{minipage} & \begin{minipage}[b]{\linewidth}\raggedright
eval after
\end{minipage} & \begin{minipage}[b]{\linewidth}\raggedright
train mode
\end{minipage} & \begin{minipage}[b]{\linewidth}\raggedright
gap
\end{minipage} \\
\midrule\noalign{}
\endhead
\bottomrule\noalign{}
\endlastfoot
released configuration & 1.4585 & \textbf{0.3064} & 0.3076 (unchanged) & 4.7$\times$ $\rightarrow$ 1.0$\times$ \\
corrected pipeline & 4.6034 & \textbf{0.0086} & 0.0150 (unchanged) & 302.7$\times$ $\rightarrow$ 0.6$\times$ \\
already-calibrated control & 0.1846 & \textbf{0.1811} & 0.1811 (unchanged) & 1.02$\times$ $\rightarrow$ 1.00$\times$ \\
\end{longtable}
}

The first row lands on the training-mode value we had measured independently
beforehand, which validates the procedure on a case whose answer was known. The
third row is the control: on a checkpoint whose statistics were already
correct, recalibration moves the held-out loss by under two per cent while
driving the residual mode effect from +0.0034 to exactly zero. Recalibration
is a repair, not a general performance improvement.

Two further quantities we report elsewhere are distorted by the same cause, and
we flag them here because neither is a loss and neither would be expected to
depend on a normalisation layer's stored statistics.

The effective rank of the embedding cloud (\S{}5.4) is computed downstream of the
same layer. Recalibration modifies no weight, yet it moves the
released-configuration checkpoint's rank from 11.9 to 18.6 and the corrected
checkpoint's from 16.5 to 67.8. A measurement that changes by half on a model
that did not change is measuring the statistics rather than the model.

Planning outcomes are sensitive in the same way, at a magnitude the loss does
not reveal. Recalibrating the already-calibrated checkpoint of the third row
above moved its held-out loss by under two per cent and yet changed the outcome
of seven of fifty planning episodes, because a change of 6$\times$10\textsuperscript{-7} in the running
variance is amplified by a factor of 78 before it reaches the planner's cost
function. Reporting a loss to four decimal places is not sufficient evidence
that two checkpoints will behave alike.

One practical note for anyone auditing a checkpoint rather than training one:
the count of batches accumulated into the normalisation statistics distinguishes
the two states directly. Ours record between 36,748 and 54,206 batches --- the
exponential moving average maintained during training --- while a recalibrated
checkpoint records the 124 to 224 batches of the precise-BN pass.

The consequence for checkpoint selection is worth stating separately, because it
is easy to reproduce elsewhere. Our training loop saved a ``best'' checkpoint by
validation loss. Under this artifact, that criterion does not select the best
model: it selects the epoch whose normalisation statistics happen to be best
calibrated. In the cyclic run, the saved checkpoint is precisely the epoch at
which the gap reaches 1.00$\times$.

We emphasise the scope. This is a property of our measurement of our
reimplementation, using an architecture the released configuration specifies. We
make no claim about the original authors' training procedure, which may
recalibrate, evaluate differently, or never encounter drift of this size.

\subsection{4.4 Training under the released and the corrected configuration}\label{training-under-the-released-and-the-corrected-configuration}

With the measurement repaired, the training results can be read directly. Both
runs below use the same data, clip index, learning rate, regularisation weight,
context length, schedule (none), epoch count and seed; they differ only in the
four pipeline corrections of \S{}3.2.

\textbf{The released configuration, as reimplemented, plateaus.} Its training loss
reaches approximately 0.30 within the first epoch and stays there --- the median
per-epoch value moves from 0.292 to 0.304 over ten epochs, a drift of under 4\%,
with a fitted slope of +0.0004 per epoch. Recalibrated held-out loss: \textbf{0.3064}.
This is not instability; it is a floor.

The floor has a straightforward cause, and it is the deviation quantified in
\S{}3.2. Under our original clip loader the predictor received a single
sub-sampled action and was asked to explain a five-step displacement, with the
remaining four actions unobserved. Measured on the released dataset, the action
supplied implies a displacement wrong by a median of 25.59 units against a
typical per-block movement of 13.3. From the predictor's perspective the
majority of its target's variance was unexplainable from its input, and a
conditional mean is the best available fit.

\textbf{The corrected configuration converges.} With actions gathered densely and the
input normalisations applied, the training loss descends monotonically at every
epoch --- 0.0412, 0.0268, 0.0226, 0.0205, 0.0186, 0.0176, 0.0166, 0.0159, 0.0148,
0.0146 --- a 65\% reduction with no oscillation, and the lowest value at the final
epoch. Recalibrated held-out loss: \textbf{0.0086}, against 0.3064 for the released
configuration. The corrected pipeline is \textbf{36 times better} on the same
held-out clips.

The resulting model predicts substantially better than a frozen-world baseline.
Given the dense action sequence it was trained on, its one-step error is
\textbf{0.068} relative to that baseline; given the displacement-matched
constant-action encoding that a planner is able to emit, \textbf{0.116}. The gap
between those two figures --- a factor of 1.7 --- is what a planner gives up by
being unable to vary its action within a frameskip block, and we report it as a
stated limitation rather than a surprise. Removing the action normalisation
alone moves the second figure to 0.337, confirming empirically a deviation we
found only by reading the reference source.

Two remarks on scope. Our runs use ten epochs, following the paper's appendix,
while the repository configuration specifies one hundred (Table 1); the
convergence reported here is therefore convergence within the paper's stated
budget, not a claim about the asymptote. And all figures are from a single seed.

Finally, the regulariser behaves as the original describes in both runs; the
embedding-spread series is reported in \S{}4.1.

\subsection{4.5 Planning at both published goal offsets}\label{planning-at-both-published-goal-offsets}

The original reports approximately 87\% of goals reached on TwoRoom under
cross-entropy-method planning over the learned model \citep[Fig.~6]{maes2026lewm}. Under our
protocol at the repository's evaluation goal offset of 25 steps, our corrected
reproduction reaches \textbf{47 of 50 = 94.0\%} (non-trivial 45 of 48 = 93.8\%). The
95\% Wilson interval is {[}83.8\%, 97.9\%{]} and contains the reported figure; a
one-sample test against 0.87 gives p = 0.203.

Two comparisons place that number, and both use the identical fifty episodes.

The authors' own released checkpoint, driven through our harness with only the
weights changed, reaches \textbf{42 of 50 = 84.0\%}. Our checkpoint is higher, but the
matched-pair test gives 5 improvements against 0 reversals, \textbf{p = 0.0625} --- a
difference not established at this sample size. What the comparison does establish is that our evaluation protocol is faithful
under the reading that reproduces: it recovers the reported result from the
reported weights under the repository's evaluation configuration, though not
under the protocol the paper's appendix describes (\S{}4.2).

Our own earlier checkpoint, trained before the pipeline corrections of \S{}3.2,
reaches \textbf{39 of 50 = 78.0\%}. Against it, the corrected checkpoint improves 11
episodes and loses 3, \textbf{p = 0.0574} --- again higher but not established at
n = 50. We note that both checkpoints were normalisation-recalibrated before this comparison (\S{}4.3). The same comparison against the un-recalibrated checkpoint would confound the pipeline correction with the recalibration, and we do not report it.

One structural detail supports the reading that these are genuine model
differences rather than measurement noise. Our corrected checkpoint fails on
three episodes, and \textbf{all three are among the eight on which the authors'
checkpoint also fails}. The failures nest rather than scatter, and all three
are among the longest goals in the set (56.9, 72.1 and 96.8 units against a
median of roughly 48).

Two qualifications belong with the headline figure. First, the 87\% is measured
under the authors' episode selection, which is not published; ours is a fixed
random draw at a stated seed, and remains a listed deviation (Table 1). The like-for-like comparison is therefore 94.0\% against 84.0\% at goal offset 25
on identical episodes, not against 87\%. Second, all three of our figures come from a single
seed, and we make no claim about seed variance.

Finally, the number is specific to the goal offset and the step budget. The
repository's evaluation configuration uses a 25-frame offset and a 50-step
budget; Appendix F.1 states 100 and 150 (\S{}4.2). The choice is consequential:
the same checkpoint reaches 94.0\% at offset 25, 20.0\% at offset 100 within the
repository's budget, and 26.0\% within the paper's. Section 5.3 takes that up,
because the effect is not a simple degradation with distance --- at the longer
offset our least accurate checkpoint is the strongest planner.

\section{5 Findings beyond the reproduction}\label{findings-beyond-the-reproduction}

\subsection{5.1 A silent evaluation-domain gap}\label{a-silent-evaluation-domain-gap}

Our most transferable finding concerns not the method but the way it was
evaluated. For three paid training runs we measured planning success in a
32$\times$32-pixel fixture built to iterate cheaply on a CPU laptop, and every one of
those measurements was worthless. The fixture reproduces the environment's
layout faithfully enough that the two are difficult to tell apart by eye: same
two rooms, same dividing wall, same door, same red agent. It differs in an
inset border with corner ticks, and in a door that is narrower and higher.

Those differences are invisible to a human and decisive for a Vision
Transformer. Encoding fixture frames and measuring the distance to the nearest
real-data latent gives a median of 61.03, against a nearest-neighbour spacing
within the real data of 2.43 --- the evaluation frames sat twenty-five times
further from the training distribution than training frames sit from each
other. Every planning number produced there was an extrapolation, and all of
them came back below a random-action control.

Three properties of this failure are worth stating, because each defeated a
check we thought sufficient.

It is \textbf{invisible to a representation probe.} A linear probe recovering agent
position from the encoder's output scored $R^{2}$ 0.9916 throughout, and the probe
itself is evaluated on real frames, so it never registered the shift.

It is \textbf{fully explained by rendering style alone.} Holding the environment
dynamics fixed and changing only the renderer reproduces the anomalous result to
within noise (57.5 against the observed 56.6), so no property of the dynamics,
the planner, or the checkpoint is required to explain three runs of apparent
failure.

It is \textbf{cheap to detect once measured rather than argued.} The same
instrument --- median latent distance from evaluation frames to their nearest
training-set neighbour --- reads 61.03 on the fixture and 0.01 on the real
environment. We now run it as a precondition inside every evaluation, which
refuses to emit a success rate when the check fails.

The general lesson is that visual fidelity is not distributional fidelity, and
that a probe demonstrating a representation is good does not demonstrate that
the inputs being fed to it are in-distribution. A debugging fixture that looks
right is exactly the kind of artifact that survives review by inspection.

That lesson recurs in a different form in \S{}4.3, where a normalisation layer's
stored statistics --- not the model, and not the data --- determined a reported
loss for three training runs. In both cases a quantity we were measuring
routinely and reporting confidently was a property of the instrument rather than
of the system. In both cases the check that would have caught it was cheap, and
in neither case had we thought to run it. We return to this in \S{}6.3.

\subsection{5.2 A pre-registered effect that did not survive a change of checkpoint}\label{a-pre-registered-effect-that-did-not-survive-a-change-of-checkpoint}

A published critique of latent world models argues that scoring a plan by
Euclidean distance between latent states conflates latent proximity with
reachability, and predicts that a planner will fail disproportionately when a
goal requires moving \emph{away} from it before approaching \citep{li2026beyond}. TwoRoom
provides a clean test: goals in the opposite room require routing through a
single door, and goals in the same room do not. We designed and pre-registered
an experiment to measure it, obtained a large and highly significant effect, and
then found that the effect does not survive either a change in how the model is
driven or a change of checkpoint. We report all three arms.

\subsubsection{The design}\label{the-design}

The obvious confound is distance: cross-wall goals are farther on average. We
therefore built a matched-pair design. From the 6,056 episodes long enough to
supply a goal at our longer offset, we performed one-to-one caliper matching on
start-to-goal distance, obtaining \textbf{110 matched pairs}. Matching was possible at every distance band
up to approximately 203 units; beyond that no same-room counterpart exists,
since a same-room goal cannot exceed the diagonal of a half-arena, so 22 of
4,373 cross-wall candidates fall outside the claim's scope. Matching quality far
exceeded the caliper: the median within-pair distance difference was 0.02 units
against a caliper of 6, the worst was 0.35, and the two arms' median distances
were identical at 124.8. A simulated power analysis on the matched-pair test
gave 82\% power for the 20-point difference we assumed. The design, the decision
rule for every outcome, and the exact list of 220 episodes were committed to the
repository before any of them was evaluated.

Two further explanations were eliminated after the fact. First, reachability: a
door-routing oracle with access to true positions, planning no further ahead
than the real dynamics allow, reached the goal in \textbf{110 of 110 episodes in both
arms} within the step budget, so the achievable ceiling is 100\% for both
geometries. Second, direction: same-room goals in this set are predominantly
vertical and cross-wall goals predominantly horizontal, but the oracle clears
both at 100\%, so orientation cannot account for a difference. A residual
path-length effect was bounded analytically: deleting the quarter of cross-wall
episodes with the longest geometric requirement and counting every one of them
as a length-caused failure still leaves a 27-point difference.

One property of the underlying population is worth stating, because \S{}4.2
establishes it only after this experiment was designed. Episodes long enough to
supply a goal 100 frames ahead are exactly the 6,056 that ran to the dataset's
length cap --- the episodes in which the data-collection policy timed out. The
matched set is therefore drawn entirely from that policy's failures, and every
goal in it is a position the policy had not reached. This does not affect the
matching, the power calculation or the oracle ceiling, all of which are computed
on the episodes as selected. It does mean that what these three arms measure is
not a task the data-collection policy solved.

\subsubsection{The three arms}\label{the-three-arms}

{\def\LTcaptype{none} 
\begin{longtable}[]{@{}
  >{\raggedright\arraybackslash}p{(\linewidth - 8\tabcolsep) * \real{0.2000}}
  >{\raggedright\arraybackslash}p{(\linewidth - 8\tabcolsep) * \real{0.2000}}
  >{\raggedright\arraybackslash}p{(\linewidth - 8\tabcolsep) * \real{0.2000}}
  >{\raggedright\arraybackslash}p{(\linewidth - 8\tabcolsep) * \real{0.2000}}
  >{\raggedright\arraybackslash}p{(\linewidth - 8\tabcolsep) * \real{0.2000}}@{}}
\toprule\noalign{}
\begin{minipage}[b]{\linewidth}\raggedright
driving regime
\end{minipage} & \begin{minipage}[b]{\linewidth}\raggedright
same-room
\end{minipage} & \begin{minipage}[b]{\linewidth}\raggedright
cross-wall
\end{minipage} & \begin{minipage}[b]{\linewidth}\raggedright
difference
\end{minipage} & \begin{minipage}[b]{\linewidth}\raggedright
exact p
\end{minipage} \\
\midrule\noalign{}
\endhead
\bottomrule\noalign{}
\endlastfoot
pre-correction checkpoint, as measured & 79.1\% & 40.0\% & \textbf{+39.1} {[}+27.2, +51.0{]} & 3.4 $\times$ 10\textsuperscript{-8} \\
pre-correction checkpoint, action-scale corrected & 74.5\% & 61.8\% & +12.7 {[}+0.5, +24.9{]} & 0.054 \\
corrected checkpoint & 14.5\% & 20.9\% & $-$6.4 {[}$-$16.4, +3.7{]} & 0.248 \\
\end{longtable}
}

The first row is the pre-registered primary test and stands as registered: on
that checkpoint, with distance matched pair-by-pair and reachability verified at
100\% for both geometries, the planner reached 79.1\% of same-room goals and 40.0\%
of cross-wall goals, a difference of 39.1 points at p = 3.4 $\times$ 10\textsuperscript{-8}.

The second row applies the action convention of \S{}3.2 to the same checkpoint,
supplying the planner's action as the block sum it was trained on rather than
as a single held step --- the same correction measured on the authors' weights in
\S{}4.2. The difference falls to +12.7 points and the test becomes
marginal. The third row uses the corrected checkpoint of \S{}4.4. The point estimate
is now negative, and the test is not significant.

The estimate therefore falls monotonically across the three arms and crosses
zero. We are careful about what the third arm establishes. It \textbf{rules out} an
effect as large as the +20 points the study was designed to detect: the
confidence interval excludes it. It does \textbf{not} establish a reversal --- at
p = 0.248 the honest reading is no detectable difference. And a confound must be
named: overall success across the three arms is 59.5\%, 68.2\% and 17.7\%, so they
are not compared at matched performance, and the low rate in the third arm could
in principle mask a real difference. The random-action control reached 0 of 110
in both geometries in every arm; being at the floor, it is uninformative about
whether the episodes differ in intrinsic difficulty, and we do not use it as
evidence that they do not.

\subsubsection{What we take from it}\label{what-we-take-from-it}

We do not claim there is no geometric effect on this task, and we do not claim
the published critique is wrong. What our three arms support is narrower and, we
think, more useful: \textbf{a large, well-powered, pre-registered, confound-controlled
effect measured on one reproduction checkpoint survived neither a change in how
that checkpoint was driven nor a change of checkpoint.} Everything a careful
reader would ask of the first row --- matching, power, a committed episode list, an
oracle ceiling, an analytic bound on the residual confound --- was in place, and
none of it was sufficient to make the number a property of the method rather
than of the checkpoint.

The implication for reproduction work is direct. A reproduction that establishes
a mechanism-level effect on its own trained checkpoint has established it for
that checkpoint. Whether it generalises is a separate question requiring
separate checkpoints, and in our case the answer was no. We report the
pre-registered result because we pre-registered it, and we report the other two
arms because they are what the result turned out to mean.

\subsection{5.3 One-step accuracy does not predict long-horizon planning}\label{one-step-accuracy-does-not-predict-long-horizon-planning}

A world model is trained to predict, and used to plan. It is natural to treat
prediction error as a proxy for planning competence. Across three checkpoints
spanning a sevenfold range in one-step prediction error, we find that the proxy
holds at short horizons and fails at long ones --- and that the two most accurate
models plan \textbf{worse than a random-action control} at the longer horizon.

Table 3 gives the comparison. All three checkpoints are evaluated under one planning protocol on identical episodes; one-step error is reported relative to a frozen-world baseline, so a value below 1 means the model predicts better than assuming nothing moves. The one-step figures are separate in-domain measurements on real validation clips, each taken under the constant-action encoding a planner actually emits, and each committed: \texttt{runs\_archive/verified/driving\_spec\_phase2\_recal.txt} for the corrected checkpoint, \texttt{runs\_archive/verified/repeat\_encoding\_authors.txt} for the authors' released weights, and \texttt{runs\_archive/verified/check\_a\_step0\_run2.txt} for the pre-correction checkpoint. They come from three different instruments at different sample draws, so the third decimal is not comparable across rows; the ordering, which is all the argument uses, is unaffected. Because the released material specifies two different
step budgets (\S{}4.2), we report the long-horizon task under both.

At goal offset 25 the ordering is monotone: as one-step error falls from 0.829 to 0.410 to 0.116, success rises from 78.0\% to 84.0\% to 94.0\% (Figure 2a, upper line). Prediction accuracy behaves exactly as the proxy assumption expects.

At goal offset 100 the ordering does not hold at all. Under the repository's
50-step budget, success runs 54.0\%, 12.0\% and 20.0\% over the same three
checkpoints. The \textbf{least} accurate model is by a wide margin the best
long-horizon planner: against it, the authors' checkpoint loses 23 episodes and
gains 2 (p = 1.9$\times$10\textsuperscript{-5}), and our corrected checkpoint loses 18 and gains 1
(p = 7.6$\times$10\textsuperscript{-5}). The two more accurate checkpoints are not distinguishable from
each other (p = 0.29).

The step budget separates the two failure modes further, and does so in the
direction an overshoot account predicts. Raising the budget at goal offset 100
from 50 steps to the 150 the paper's appendix specifies lifts the least accurate
checkpoint from 54.0\% to \textbf{80.0\%} --- twenty-six points --- while moving the
authors' checkpoint two points and our corrected checkpoint six. Extra time
rescues a planner that was running out of it, and does almost nothing for
planners that are travelling in the wrong direction. At the longer budget the
dissociation is sharper rather than weaker: one-step errors of 0.829, 0.410 and
0.116 map to 80.0\%, 14.0\% and 26.0\%.

The failure mode is overshoot rather than stalling, and it is visible without
any modelling. Under the 50-step budget the random-action control finishes a
mean of 111.1 units from the goal, while the two accurate checkpoints finish at 116.6 and 122.5 units --- \textbf{farther away than random} (Figure 2b) --- with individual final
distances of 140 to 193 units in an arena roughly 192 units across. The least
accurate checkpoint finishes at 40.5 units. Neither accurate model is incapable
of long goals: our corrected checkpoint reached a 173-unit goal in 45 of its 50
allotted steps. They systematically travel too far.

The longer budget partly softens that picture, and the way it softens is
informative. Our corrected checkpoint draws level with its own random control
(108.8 units against 108.1) while the authors' checkpoint remains farther out
(122.7 against 108.1). The additional steps let the more accurate model stop
diverging; they do not let it arrive.

That the pattern holds for the \textbf{authors' own released weights}, and most
strongly there, matters for how it should be read. It is not an artifact of our
reimplementation, our pipeline corrections, or our recalibration procedure. It
is a property of this task, this planner and this class of model.

We are careful about mechanism. A plausible account is that a more accurate
model produces a sharper cost landscape, so the optimiser commits to
near-maximal actions, which a terminal-cost objective does not penalise until
the horizon ends; a weaker model yields a flatter landscape and more moderate
actions. The mean-final-distance column and the budget response are both
consistent with this, but we have not tested it, and we do not claim it.

A confound must also be stated plainly. The two overshooting checkpoints both
use a three-frame context; the cautious one uses a single frame. Context length
is therefore an alternative explanation to prediction accuracy, and three
checkpoints cannot separate the two. Distinguishing them would require training
matched checkpoints that vary one factor at a time, which our compute budget did
not allow.

The practical implication stands regardless of mechanism. \textbf{Selecting a world
model by held-out one-step prediction error is not a reliable way to select a
world model for long-horizon planning}, and on this task at this horizon it
would have selected the worst of three available options under either budget.
Reporting a single planning number without its horizon is correspondingly
misleading: across these three checkpoints the goal offset alone moves success
between 12\% and 54\% at the repository's budget, and between 14\% and 80\% at the
paper's.

\medskip\noindent
\textbf{Table 3: One-step prediction error against planning success at two goal
horizons and two step budgets.} All figures from 50 episodes per cell,
identical across checkpoints, under one protocol. One-step error is relative
to a frozen-world baseline (below 1 = better than assuming no motion). Mean
final distance is at offset 100 under the 50-step budget, where the
random-action control finishes at 111.1 units.
\medskip

{\def\LTcaptype{none} 
\begin{longtable}[]{@{}
  >{\raggedright\arraybackslash}p{(\linewidth - 12\tabcolsep) * \real{0.1429}}
  >{\raggedright\arraybackslash}p{(\linewidth - 12\tabcolsep) * \real{0.1429}}
  >{\raggedright\arraybackslash}p{(\linewidth - 12\tabcolsep) * \real{0.1429}}
  >{\raggedright\arraybackslash}p{(\linewidth - 12\tabcolsep) * \real{0.1429}}
  >{\raggedright\arraybackslash}p{(\linewidth - 12\tabcolsep) * \real{0.1429}}
  >{\raggedright\arraybackslash}p{(\linewidth - 12\tabcolsep) * \real{0.1429}}
  >{\raggedright\arraybackslash}p{(\linewidth - 12\tabcolsep) * \real{0.1429}}@{}}
\toprule\noalign{}
\begin{minipage}[b]{\linewidth}\raggedright
checkpoint
\end{minipage} & \begin{minipage}[b]{\linewidth}\raggedright
one-step error
\end{minipage} & \begin{minipage}[b]{\linewidth}\raggedright
offset 25, budget 50
\end{minipage} & \begin{minipage}[b]{\linewidth}\raggedright
offset 100, budget 50
\end{minipage} & \begin{minipage}[b]{\linewidth}\raggedright
offset 100, budget 150
\end{minipage} & \begin{minipage}[b]{\linewidth}\raggedright
mean final dist. @100/50
\end{minipage} & \begin{minipage}[b]{\linewidth}\raggedright
context frames
\end{minipage} \\
\midrule\noalign{}
\endhead
\bottomrule\noalign{}
\endlastfoot
our pre-correction checkpoint & 0.829 & 78.0\% & \textbf{54.0\%} & \textbf{80.0\%} & \textbf{40.5} & 1 \\
authors' released & 0.410 & 84.0\% & \textbf{12.0\%} & \textbf{14.0\%} & 122.5 & 3 \\
our corrected checkpoint & 0.116 & \textbf{94.0\%} & 20.0\% & 26.0\% & 116.6 & 3 \\
random-action control & --- & 18.0\% & 0.0\% & 2.0\% & 111.1 & --- \\
\end{longtable}
}

\begin{center}
\includegraphics[width=\textwidth]{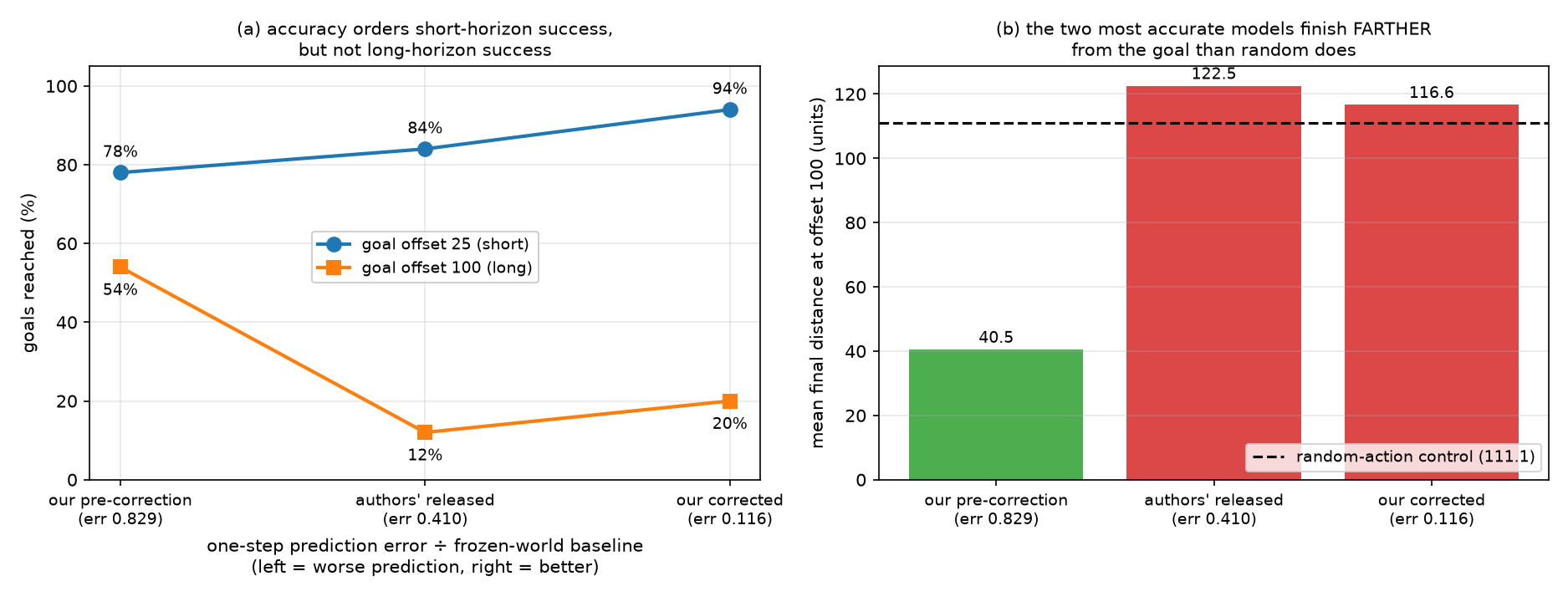}
\end{center}

\medskip\noindent
\textbf{Figure 2: Prediction accuracy orders short-horizon planning success and fails to order long-horizon planning success.} \textbf{(a)} Goals reached against
one-step prediction error, with the axis running from worse to better
prediction. At goal offset 25 the relationship is monotone; at offset 100 it
is not, and the least accurate checkpoint is the strongest planner. \textbf{(b)}
Mean final distance at offset 100. Bars above the dashed line finish farther
from the goal than a random-action policy does; the two most accurate
checkpoints both do.
\medskip

\subsection{5.4 What the corrected pipeline does to the representation}\label{what-the-corrected-pipeline-does-to-the-representation}

The pipeline corrections of \S{}3.2 change the predictor's task substantially, and
it is natural to ask what they do to the representation the encoder learns. We
compared the two encoders --- one trained under the released configuration, one
under the corrected pipeline --- on 4,000 identical held-out frames, with each
encoder receiving the pixel convention it was trained with.

Three of the four measurements are unchanged. Position is decodable at $R^{2}$ 0.9977
against 0.9971 by a linear probe, and at 0.9994 by a two-layer network for
both. The summed action executed between two frames is decodable from the
pair of embeddings at 0.9290 against 0.9132. On the information a planner needs,
the two encoders are equivalent.

The fourth measurement is not. The \textbf{effective rank} of the embedding cloud ---
the participation ratio of its covariance spectrum, which counts how many
dimensions the representation actually occupies --- rises from \textbf{18.6 to 67.8 of
192}. The corrected pipeline produces a representation spread over roughly
three and a half times more dimensions while carrying the same position and
action information.

We tested one explanation and it did not survive. Our hypothesis was that the
extra dimensions carry dynamics-relevant structure, purchased at the cost of
some of the linear structure that a position probe reads: under the released
configuration the predictor received one sub-sampled action to explain a
five-step displacement (\S{}3.2), so it could not usefully constrain the encoder,
leaving it free to become a nearly pure position code. That hypothesis predicts
that the corrected encoder should make actions \emph{more} linearly decodable. It
does not --- action decodability is unchanged to within 0.016. We therefore report
the rank difference as an observation and offer no account of what the
additional dimensions encode.

Two cautions apply to reading the rank difference at all. First, the two
encoders differ in three respects rather than one: the action aggregation, the
pixel convention --- raw \texttt{{[}0,1{]}} against ImageNet-normalised --- and the
action-encoder width (\S{}3.2). Two checkpoints cannot attribute a difference to
one of three simultaneous changes, and we do not. Second, effective rank is
computed downstream of the normalisation layer of \S{}4.3, and is therefore partly
a measurement of that layer's stored statistics rather than of the encoder; the
paragraph below shows it moving by more than half on a checkpoint whose weights
never changed.

The original offers an explanation for its TwoRoom result that this measurement
speaks to directly. Its discussion attributes the environment's weak planning
performance to the low diversity and low intrinsic dimensionality of the
dataset, which it argues makes it difficult for the encoder to match the
isotropic Gaussian prior the regulariser enforces in a high-dimensional latent space \citep[\S4.2]{maes2026lewm}. Our reference-faithful encoder occupies 18.6 of 192 available
dimensions, which is consistent with that account and puts a number on it. What
we add is that the occupancy is not fixed by the environment: correcting the
action pipeline raises it to 67.8 while leaving the decodable content unchanged,
so at least part of the shortfall against the prior is a property of the
training pipeline rather than of the dataset.

We record one methodological point, because it changed our own answer twice.
Both measurements above were first taken before the normalisation repair of
\S{}4.3, and both were wrong: the corrected encoder then appeared to have an
effective rank of 16.5 rather than 67.8, and appeared to lose action
decodability (0.8733 rather than 0.9132). The second error was subtler, and we
caught it only on re-verification: having repaired one checkpoint, we compared
it against the other unrepaired. That understates the reference-faithful
encoder's own rank as 11.9 rather than 18.6, and inflates the reported
difference from 49 dimensions to 56. The position figures were unaffected,
because a ridge probe
standardises its inputs and is scale-invariant. Any measurement on a latent space
whose scale a normalisation layer controls should be taken after verifying that
layer, and probe-style measurements are precisely the ones that will not warn
you.

\subsection{5.5 Scoring geometry within the training distribution}\label{scoring-geometry-within-the-training-distribution}

A published critique of latent world models argues that scoring plans by
Euclidean distance in latent space conflates latent proximity with reachability
\citep{li2026beyond}. \S{}5.2 tests the behavioural prediction that follows from it.
Here we report the representation-level measurement, which is narrower and
points the other way.

Sampling pairs of real frames at matched physical distance and comparing their
latent separation, embeddings of positions in \emph{different} rooms are separated by
\textbf{1.79 times} as much as embeddings of positions in the same room at the same
physical distance. The wall is represented: at equal Euclidean distance in the
arena, the latent space places cross-wall pairs farther apart, which is the
direction the scoring function would need in order to prefer routing. The strong
form of the critique --- that the latent geometry is blind to the obstacle --- is
therefore not supported for this encoder.

Two qualifications. This measurement is from a single checkpoint, taken before
the pipeline corrections, and was not repeated afterwards. And given \S{}5.2, where
the behavioural effect this measurement was originally offered to explain did
not survive a change of checkpoint, we do not present the 1.79$\times$ figure as
support for any behavioural claim. It is a property of one encoder's latent
geometry, reported as such.

\section{6 Discussion}\label{discussion}

\subsection{6.1 What was easy}\label{what-was-easy}

The released environment installs from PyPI and runs without modification. The
released checkpoint downloads from the model hub and, once the architecture is
reconstructed from its configuration, loads with strict key matching. The
representation result is easy to obtain and robust: a position probe reaches
$R^{2}$ 0.99 within a single epoch under every pipeline configuration we tried, and
never degraded thereafter. The anti-collapse regulariser behaves exactly as
described, in every run, without tuning.

\subsection{6.2 What was difficult}\label{what-was-difficult}

\textbf{The four deviations that mattered were invisible in configuration files.}
Dense action gathering, the programmatic action-encoder width, ImageNet pixel
normalisation and action z-scoring are all determined in code, and a reproducer
following the released configuration alone obtains a silently broken model.
Three of the four we found only by reading the reference's data loader and
training script line by line, after a measurement told us something was wrong;
the fourth we found only because their released weights refused to behave.

\textbf{The evaluation-domain gap survived a 0.99 probe for three paid runs (\S{}5.1),
and a normalisation artifact concealed the training result for as long
(\S{}4.3).} Both were failures of instrumentation rather than of the method under
study, and in both cases we reported the wrong conclusion confidently before
finding the cause.

\textbf{Ten thousand NaN actions sit at the end of each episode in the released
dataset.} Our original loader read one action in five and stepped over them by
luck; the corrected loader reads all of them, and would have produced NaN
gradients from the first affected batch. The reference drops NaN rows before
computing normalisation statistics --- one line we read and did not implement.

\textbf{Our own checks failed more often than the runs did.} Of the gate failures we
investigated, more were caused by defects in the gate than by defects in the
run: a physics assertion measured on normalised rather than raw actions, a
collapse test invalid at the batch size it ran at, a log reader that silently
took the wrong file, and a driving-spec diagnostic that scored a working model
as unusable because it supplied displacement-mismatched actions. We report this
because a reproduction paper that documents only the subject's failures is not
being straight about where the effort goes.

\subsection{6.3 Recommendations}\label{recommendations}

The following are addressed to authors releasing work and to reproducers
attempting it. Each is drawn from a specific failure above.

\textbf{1. Publish the data convention, not only the configuration.} Action
aggregation and input normalisation determined every result in this paper, and
neither appears in any released configuration file. A short section of the
README stating how actions are aggregated across a frameskip block, and what
normalisation is applied to inputs, would have saved us several days and one
paid training run. This is the single highest-value change available to authors
of the work we reproduced.

\textbf{2. Check fidelity against source, not configuration.} Our audit compared 40 pipeline elements against the reference implementation's source with
file-and-line citations, and marked each as matching, deviating, or unverified
(Table 1). Every one of the four expensive deviations lived in code
that no configuration file mentions. An audit against configurations would have
found none of them, and we had performed exactly such an audit twice before,
concluding both times that we matched.

\textbf{3. Assert data contracts against physics, not shapes.} The environment we
studied is deterministic, so displacement across a block equals speed times the
summed actions of that block. That identity is one line of code, and it
detects the action-aggregation deviation immediately and unambiguously. A
shape check does not: the incorrect array had the correct rank, the correct
dtype, and plausible magnitudes. Where a dataset admits an exact invariant,
assert the invariant.

\textbf{4. Differential-test against a released artifact.} Running the authors'
released checkpoint through our own evaluation harness (\S{}4.2) established that
our protocol reproduces the reported result, which no amount of internal
consistency checking could have established. It also caught a convention error
on our side, because their weights refused to behave when driven incorrectly. If
a reproduction target releases weights, running them through your harness before
trusting your own numbers is the highest-information check available.

\textbf{5. Verify that evaluation mode measures the model, not the normalisation.}
This is the recommendation we most wish we had received. Where a network
contains batch normalisation, compare its layers' running variance against the
scale of the activations reaching them. In our checkpoints the projector's
running variance was of order 10\textsuperscript{-4}, so evaluation mode divided by 0.014 and
amplified any staleness in the stored statistics by a factor of 72 or more; in
the authors' released checkpoint the same layer holds 0.0172 and amplifies by
7.6, and its evaluation mode is faithful. Two symptoms are worth watching for:
a validation loss that oscillates while the training loss does not, and a gap
between the two that shrinks as the learning rate falls. Recalibrating the
running statistics --- a few hundred forward passes in training mode with no
gradient updates --- is a cheap repair, and on a checkpoint whose statistics are
already correct it is a verified no-op (\S{}4.3).

A corollary concerns checkpoint selection. Under this artifact, saving the
checkpoint with the best validation loss does not select the best model: it
selects the epoch whose normalisation statistics happen to be best calibrated.
In our cyclic run, the saved checkpoint is precisely the epoch at which the gap
reaches 1.00$\times$, and we spent some time believing that epoch's weights were
special.

\subsection{6.4 Communication with the original authors}\label{communication-with-the-original-authors}

We wrote to the corresponding author on 31 July 2026, reporting the two
undocumented preprocessing steps of \S{}3.2 and asking four questions. As of
submission we have received no response.

We record them here alongside what the released artifacts show, because three of
the four can be settled from those artifacts alone. Saying so makes the
correspondence a request for confirmation rather than for information, and it
lets anyone reproducing this work resolve the same ambiguities without waiting
for an answer.

{\def\LTcaptype{none} 
\begin{longtable}[]{@{}
  >{\raggedright\arraybackslash}p{(\linewidth - 2\tabcolsep) * \real{0.5000}}
  >{\raggedright\arraybackslash}p{(\linewidth - 2\tabcolsep) * \real{0.5000}}@{}}
\toprule\noalign{}
\begin{minipage}[b]{\linewidth}\raggedright
question
\end{minipage} & \begin{minipage}[b]{\linewidth}\raggedright
what the released artifacts show
\end{minipage} \\
\midrule\noalign{}
\endhead
\bottomrule\noalign{}
\endlastfoot
Which \texttt{history\_size} is operative for TwoRoom? Appendix D states 1; the repository configuration specifies 3. & \textbf{3.} The released checkpoint's \texttt{predictor.pos\_embedding} has shape (1, 3, 192), so 3 is the value that produced it. \\
Are dense action gathering and ImageNet normalisation intended as we describe them, given that neither appears in any configuration file? & \textbf{Yes} to both, from \texttt{buffer.py} and \texttt{utils.py} respectively (\S{}3.2), and corroborated by the released weights. A reader following the configuration alone would implement neither. \\
Which goal offset and step budget produced the reported figure? Appendix F.1 states a 150-step budget with the goal sampled 100 steps ahead; the released evaluation configuration uses 50 and 25. & \textbf{The configuration's values.} Their own released checkpoint scores 84.0\% under 50 and 25, and 14.0\% under 150 and 100 (\S{}4.2). Only one of the two published protocols reproduces the reported result; which was intended is theirs to say. \\
Which learning rate and schedule produced the released checkpoint? & \textbf{Not recoverable from the released artifacts.} This is the one question we cannot answer for ourselves, and it bears directly on \S{}4.4. \\
\end{longtable}
}

The last of these is the one we most want answered. Our reimplementation
produces a predictor that beats a frozen-world baseline only when the learning
rate falls to about 10\textsuperscript{-6} or below, whereas the released configuration specifies
a constant 5 $\times$ 10\textsuperscript{-5} (\S{}4.4). If the released checkpoint was trained under a
schedule that does not appear in the configuration, that single fact would
account for our central negative result.

A fifth conflict --- Appendix E's ten training epochs against the repository's
\texttt{max\_epochs:\ 100} --- is tabulated in Table 1 and bears on \S{}4.4; we did not raise
it in correspondence.

\section{7 Limitations}\label{limitations}

We list the constraints on our results in roughly descending order of how much
they should change a reader's confidence.

\textbf{A single seed throughout.} Every training run, every planning evaluation and
every probe in this paper uses one seed. We make no estimate of seed variance,
and none of our differences should be read as robust to reinitialisation. This
matters most for the planning figures in \S{}4.5, where the differences we report
between checkpoints (94.0\% against 84.0\% at goal offset 25, p = 0.0625; 94.0\% against 78.0\%,
p = 0.0574) are already not established at our sample size of fifty episodes.
Our evaluations are, however, deterministic: re-running a planning evaluation
from the committed commit reproduces its per-episode outcomes exactly, so the
variance we have not measured is between-seed and not within-run.

\textbf{Ten epochs, not one hundred.} We follow the paper's appendix, which states
ten epochs; the released repository configuration specifies \texttt{max\_epochs:\ 100}
(Table 1). Our convergence result (\S{}4.4) is therefore convergence within the
budget the paper states, and says nothing about the asymptote. At four rented
GPU-runs of roughly six dollars each, a hundred-epoch run was outside our
budget, and we note that this constraint is itself a reproducibility finding: a
reader with our resources cannot test the repository's own configuration.

\textbf{Our results characterise our checkpoints, not the method.} This is the
strongest lesson of \S{}5.2. A same-room planning advantage that was pre-registered,
distance-matched, verified reachable by an oracle, and significant at
p = 3.4 $\times$ 10\textsuperscript{-8} on one checkpoint fell to +12.7 points under a change of action
scaling and to $-$6.4 points on a different checkpoint. Effect sizes measured on a
single reproduction checkpoint should not be read as properties of the method,
and we no longer read our own that way. The latent-geometry measurement of \S{}5.5 in particular was made on one
checkpoint only, before the pipeline correction, and was not repeated after it.

\textbf{The mechanism of the long-horizon reversal is untested.} Our principal
finding (\S{}5.3) is that one-step prediction accuracy orders short-horizon
planning success and fails to order long-horizon planning success, with the two
most accurate checkpoints finishing farther from the goal than a random-action
control. We offer a candidate explanation --- that a sharper cost landscape leads
the optimiser to commit to near-maximal actions that a terminal-cost objective
does not penalise until the horizon ends --- but we have not tested it, and a
confound remains: the two overshooting checkpoints use a three-frame context and
the cautious one uses a single frame. Three checkpoints cannot separate
prediction accuracy from context length. Distinguishing them requires training
matched checkpoints varying one factor at a time.

\textbf{An independent reimplementation, with a documented but non-empty deviation
set.} We did not rerun the authors' code; we reimplemented from the released
code and paper, which is what surfaced the four undocumented pipeline
differences of \S{}3.2. Our audit matched 24 of 40 elements against reference source, and the remainder are listed in Table 1. Two deviations
deserve individual mention. Our runs use full 32-bit precision where the
reference specifies bfloat16; the direction favours numerical accuracy and
cannot explain a failure to converge, but it is a difference. And the reference
computes action normalisation statistics on a two-wide raw column and applies
them to a ten-wide clip array; the broadcast it relies on is not visible in the
source we read, so we tile the two-wide statistics across the five concatenated
actions. That is our interpretation, and if it is wrong, \S{}4.4's figures move.

\textbf{Episode selection differs from the original's, which is not published.} Our
episodes are a fixed random draw at a stated seed, with the start at the first
frame of an episode and the goal a fixed number of steps later. The authors'
selection is not described in enough detail to reproduce. Consequently the
like-for-like comparison in \S{}4.5 is our checkpoint against the authors' released
checkpoint on identical episodes at goal offset 25 (94.0\% against 84.0\%), not
against the reported 87\%, which was measured under a selection we cannot replicate.

\textbf{Recalibration data overlaps the evaluation episodes.} The BatchNorm
recalibration of \S{}4.3 accumulates statistics over training clips drawn from the
same dataset as the planning episodes. Precise-BN sets normalisation statistics
only and updates no weight, so it cannot memorise episode content, and our
control run shows it is a no-op on a checkpoint whose statistics are already
correct. We nonetheless record the overlap rather than leave a reader to find
it.

\textbf{A known defect in the committed evaluation reports.} Until we corrected it,
the deviations block printed at the end of each planner report used hardcoded
default values rather than the run's actual parameters, so reports from runs at
a non-default goal offset, from the authors' checkpoint, or over the committed
episode set misdescribe themselves in that block. The measurements in those reports are unaffected --- the header of each report records the protocol that was \emph{requested}, accurately --- and the repository records which reports predate the fix. The header is not a complete description of a run. Where the data constrain a requested instruction, the protocol actually realised can be narrower than the one the header names: initial-state sampling at a goal offset of 100 is requested and recorded, and has no effect (\S{}4.2). And where two runs differ only in a driving convention the header does not print, the \texttt{spec\_as\_used.json} committed beside the report is what distinguishes them.

\textbf{Scope.} All results concern the TwoRoom diagnostic environment. We make no
claim about the original's embodied or zero-shot results, about its other
environments, or about the method's behaviour at scales other than the
18.03M-parameter configuration studied here.

\section{8 Conclusion}\label{conclusion}

The three claims we set out to test (\S{}2) resolve as follows. The representation
claim reproduces directly and easily. The planning claim reproduces under one of
the two evaluation protocols the released material publishes --- 94.0\% at the
repository's goal offset against a reported \textasciitilde87\%, and 84.0\% for the authors' own
weights under that same protocol --- once four undocumented conventions are
corrected; under the protocol the paper's appendix describes, the authors' own
weights reach 14.0\%. The training claim does not reproduce from the released
configuration files alone, and does reproduce once those conventions are
supplied --- which we take to be a documentation gap rather than a defect in the
method.

What we did not anticipate is how much of the work would consist of establishing
that our own measurements meant what we thought they meant. Three training runs
appeared not to converge because a normalisation layer's stored statistics, not
the model, determined the loss we were reporting. Three runs' worth of planning
results were produced in a debugging fixture that looks correct and sits
twenty-five times outside the training distribution. A carefully pre-registered
effect of +39.1 points at p = 3.4 $\times$ 10\textsuperscript{-8} fell to $-$6.4 points on a different
checkpoint. In each case a probe or a summary statistic read exactly as it
should have while the underlying quantity was wrong, and in each case the check
that would have caught it was cheap.

The finding we expect to be most useful outside this reproduction is the
negative one. A world model that predicts one step ahead seven times more
accurately than another was not detectably better at short-horizon planning and
was decisively worse at long-horizon planning --- and this holds for the authors'
released checkpoint as well as for ours. Selecting a world model by held-out
prediction error is not a reliable way to select a world model for planning, and
on this task at the longer horizon it would have selected the worst of the three
available.

We release the reimplementation, six checkpoints --- three BatchNorm-recalibrated and the three un-recalibrated originals they were made from, so that the evaluation-mode artifact of \S{}4.3 can be checked independently --- every evaluation report,
the fidelity audit against reference source, the pre-registration, and the gate
outputs, at (github.com/joyjeet-singh/tinylab).

\nocite{*}
\bibliographystyle{tmlr}
\bibliography{paper}

\end{document}